\documentclass[default,iicol]{sn-jnl}

\jyear{2021}%

\theoremstyle{thmstyleone}%
\newtheorem{theorem}{Theorem}
\theoremstyle{thmstyletwo}%

\theoremstyle{thmstylethree}%
\usepackage{bbding} 

\begin{document}

\title[Article Title]{Video Abnormal Event Detection by Learning to Complete Visual Cloze Tests}








\author*[1]{Siqi Wang}\email{wangsiqi10c@nudt.edu.cn}
\equalcont{These authors contributed equally to this work.}

\author[1]{Guang Yu}\email{yuguangnudt@gmail.com}
\equalcont{These authors contributed equally to this work.}

\author*[1]{Zhiping Cai}\email{zpcai@nudt.edu.cn}
\author[1]{Xinwang Liu}\email{xinwangliu@nudt.edu.cn}
\author[1]{En Zhu}\email{enzhu@nudt.edu.cn}
\author[2]{Jianping Yin}\email{jpyin@dgut.edu.cn}

\affil*[1]{\orgdiv{College of Computer}, \orgname{National University of Defense Technology}, \orgaddress{\city{Changsha}, \country{China}}}

\affil[2]{\orgname{Dongguan University of Technology}, \orgaddress{\city{Dongguan}, \country{China}}}

\abstract{Although deep neural networks (DNNs) enable great progress in video abnormal event detection (VAD), existing solutions typically suffer from two issues: (1) The localization of video events cannot be both precious and comprehensive. (2) The semantics and temporal context are under-explored. To tackle those issues, we are motivated by the prevalent cloze test in education and propose a novel approach named \textit{Visual Cloze Completion} (VCC), which conducts VAD by learning to complete ``visual cloze tests'' (VCTs). Specifically, VCC first localizes each video event and encloses it into a spatio-temporal cube (STC). To achieve both precise and comprehensive localization, appearance and motion are used as complementary cues to mark the object region associated with each event. For each marked region, a normalized patch sequence is extracted from current and adjacent frames and stacked into a STC. With each patch and the patch sequence of a STC compared to a visual ``word'' and ``sentence'' respectively, we deliberately erase a certain ``word'' (patch) to yield a VCT. Then, the VCT is completed by training DNNs to infer the erased patch and its optical flow via video semantics. Meanwhile, VCC fully exploits temporal context by alternatively erasing each patch in temporal context and creating multiple VCTs. Furthermore, we propose localization-level, event-level, model-level and decision-level solutions to enhance VCC, which can further exploit VCC's potential and produce significant performance improvement gain. Extensive experiments demonstrate that VCC achieves state-of-the-art VAD performance. Our codes and results are open at \url{https://github.com/yuguangnudt/VEC_VAD/tree/VCC}.}




\maketitle

\section{Introduction}\label{sec:intro}

\begin{figure*}
	\centering
	\includegraphics[scale=1.05]{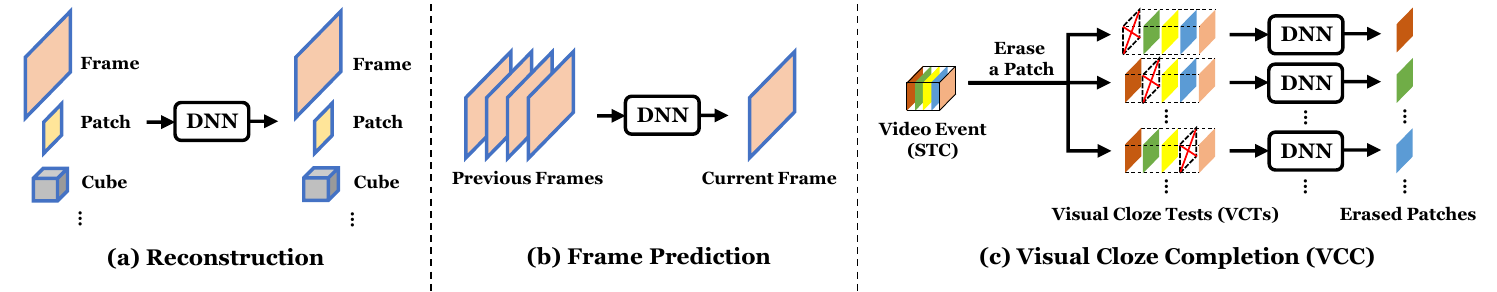}
	\caption{Learning paradigm comparison for DNN based VAD. \textbf{(a)} Reconstruction based methods train DNN to reconstruct data collected from normal training videos, e.g. video frames/patches/cubes. \textbf{(b)} Frame prediction based methods take previous frames as inputs of DNN to predict current frame. \textbf{(c)} VCC first encloses video events with STCs in a both precise and comprehensive manner. Different types of VCTs are then created by erasing the patch at different temporal positions. Afterwards, a separated DNN is trained to complete each type of VCTs, \textit{i.e.} learning to complete its erased patch or auxiliary information. Note that cubes used for reconstruction are different from STCs in VCC, as they are yielded by a relatively coarse strategy (e.g. sliding windows) and cannot enclose video events both precisely and comprehensively.}
	\label{fig:recon_pre_vec}
\end{figure*}

Video abnormal event detection (VAD), which aims to automatically detect abnormal events in surveillance videos, is an appealing topic to both academia and industry, due to its huge potential value to various safety-critical scenarios like municipal management, traffic monitoring and emergency reaction. Formally, VAD refers to detecting suspicious video events that divert significantly from the frequently-seen daily routine. With many attempts made, VAD still proves to be a challenging task, which can be ascribed to three important characteristics of abnormal events: \textbf{(1)} \textit{Scarcity}. As abnormal events typically occur at a much lower probability, it is usually difficult to collect sufficient abnormal event data. \textbf{(2)} \textit{Ambiguity}. Abnormal events refer to all events that differ from the normal observations, which makes it infeasible to enumerate all potential abnormal events for training. \textbf{(3)} \textit{Unpredictability}. It is impractical to predict the exact form of an incoming abnormal events. Due to those characteristics above, the direct modeling of abnormal events could be unrealistic. As a result, VAD usually follows the {\textit{one-class classification}} setup \cite{khan2014one}: At the training stage, abnormal events are viewed to be strictly unknown, while only normal event data (usually highly accessible) are collected. Those normal data are roughly annotated by one common label, \textit{i.e.} sub-classes in normal events are not distinguished and all of them are labeled as positive/normal. A normality model is then built with roughly labeled normal data. In the inference stage, video events that do not conform to this normality model will be viewed as abnormal. As the labels for anomalies and normal sub-classes are both absent, VAD is usually considered to be a semi-supervised task that learns with very few label, and powerful supervised learning is not directly applicable to VAD. Instead, it is usually addressed by some unsupervised or self-supervised approaches. 

In the literature, VAD solutions can be categorized into the classic VAD methods and recent DNN based VAD methods (reviewed in Sec. \ref{sec:related_work}). Classic VAD relies on hand-crafted descriptors to extract high-level features like trajectory or low-level features like texture from video events, while features are then fed into classic anomaly detection models for VAD. By contrast, DNN based VAD is inspired by DNN's success in vast vision tasks \cite{lecun2015deep}. It not only avoids complex feature engineering, but also achieves superior performance to classic VAD. Despite that DNN based VAD has achieved remarkable success and plays a dominant role in recent research, it is still faced with two prominent issues:  \textbf{(1)} Although the goal of VAD is to detect abnormal video events, \textit{existing methods for DNN based VAD actually cannot realize a both precise and comprehensive localization of video events in the first place.} The standard practice of early VAD works is to extract video events by a multi-scale sliding window with certain filtering rules \cite{xu2017detecting,sabokrou2018deep}, which cannot properly locate foreground objects and produces obviously imprecise localization. As DNNs can process high-resolution raw videos, many recent methods like \cite{liu2018future,nguyen2019anomaly,ye2019anopcn} simply overlook the event localization by learning on a per-frame basis. However, such a way is found to be vulnerable to several problems, \textit{e.g.} scale variations due to foreground depth and foreground-background imbalance \cite{liu2019exploring,zhou2019attention}. Few works  \cite{hinami2017joint,ionescu2019object} also notice such problems, and leverage an object detector pre-trained on a generic image dataset. It improves the precision but incurs another fatal ``\textit{closed world}'' problem: The pre-trained detector is unable to recognize novel foreground objects, thus leading to non-comprehensive localization. More importantly, the subjects of many abnormal events are intrinsically novel due to VAD's nature. As a result, unsatisfactory localization of video events tends to degrade later learning process. \textbf{(2)} Since a video event is essentially a high-level temporal concept, \textit{existing methods for DNN based VAD usually fail to fully exploit the video semantics and temporal context of the activity.} As illustrated by Fig. \ref{fig:recon_pre_vec}, DNN based VAD typically follows two learning paradigms (\textit{reconstruction} or \textit{frame prediction}), but both of them have their own issue: Reconstruction based methods learn to reconstruct normal events and view poorly reconstructed events as abnormal. However, simple reconstruction will drive DNNs to memorize low-level details rather than meaningful semantics \cite{larsen2016autoencoding}, and the large capacity of DNNs often enables abnormal events to be reconstructed as well \cite{Gong_2019_ICCV}. By contrast, frame prediction based methods aim to predict a normal video frame from previous frames, and poorly predicted frames are believed to contain anomalies. Prediction avoids reducing training loss by simply memorizing low-level details. Nevertheless, it typically scores each video frame only by the prediction errors of the single frame, whilst the temporal context with valuable clues of video event is not involved into the detection process. Hence, neither of two learning paradigms is a sufficiently satisfactory solution to DNN based VAD.

Unlike many recent efforts that focus on searching better DNN architectures for reconstruction or frame prediction, we are inspired by the popular cloze test in language study, and propose a new paradigm named \textit{visual cloze completion (VCC)}. As Fig. \ref{fig:recon_pre_vec} shows, the core idea of VCC is to train DNNs to complete a series of \textit{visual cloze tests (VCTs)}, which comprises of two major steps: \textbf{(1)} Extracting video events to construct VCTs. To extract video events in a both precise and comprehensive manner, we leverage appearance and motion as mutually complementary cues to locate the foreground object region associated with each video event. From each located region, a normalized patch sequence is extracted from the current and temporally adjacent frames, and then stacked into one spatio-temporal cube (STC) to enclose the video event. With each patch in STC compared to a ``word'', we can view the whole patch sequence of the STC as a ``sentence'' that describes the video event. In this way, a VCT can be constructed by erasing a certain ``word'' (patch) in the ``sentence'' (STC). \textbf{(2)} Learning to complete VCTs. Specifically, DNNs are trained to ``answer'' the VCT by inferring the erased patch, which requires DNNs to attend to the video semantics (\textit{e.g.} high-level body parts) rather than only low-level details. Meanwhile, VCC is equipped with two ensemble strategies, \textit{VCT type ensemble} and \textit{modality ensemble}: VCT type ensemble enables VCC to fully exploit the temporal context of a video event. It creates multiple types of VCTs for completion by alternatively erasing each patch in STC. In this way, each patch in the temporal context of a video event must be considered, and later an anomaly score will be computed with all possible VCTs constructed from one video event. Modality ensemble requires DNNs to infer the erased patch's optical flow, which incorporates richer motion semantics like appearance-motion correspondence. In this way, the proposed VCC parafigm is able to handle the above two issues effectively for better VAD performance.

A preliminary version of this paper is presented in \cite{yu2020cloze}. Compared with \cite{yu2020cloze}, we mainly extend the original work in terms of the aspects below: \textbf{(1)} At the level of localization, we leverage the estimated optical flow, instead of the temporal gradients used in \cite{yu2020cloze}, as motion cues for localizing video events. Optical flow enables the localization results to be more robust to noises, thus leading to less artifacts and misinterpreted video events. \textbf{(2)} At the level of video event, we design a spatially-localized strategy that aims to alleviate the scale variation problem caused by foreground depth. It divides the video frame into several non-overlapping spatial regions (\textit{a.k.a.} blocks). Video events extracted from each block are modeled separately, which enables video events with the comparable scale to be processed by DNNs. \textbf{(3)} At the level of model, we designed a new DNN architecture named spatio-temporal UNet (ST-UNet) to perform VCC. When compared with the standard UNet used in \cite{yu2020cloze}, ST-UNet synthesizes a recurrent network structure to accumulate temporal context information in STCs and produce high-level feature maps, which facilitates the proposed VCC paradigm to learn richer video semantics. \textbf{(4)} At the level of decision, we further design a mixed score metric and a score rectification strategy, which prove to be simple but highly effective strategies for performance enhancement. \textbf{(5)} As to empirical evaluation, we carry out more extensive experiments on various benchmark datasets to justify the effectiveness of VCC, and more in-depth analysis and discussion are also provided. To sum up, our main contributions are summarized below:

\begin{itemize}
	
	\item We are the first to explicitly clarify the necessity of a both precise and comprehensive video event localization, and we propose to leverage both appearance and motion as mutually complementary cues for video event extraction, which overcomes the ``closed-world'' problem and lays a firm foundation for VAD in the first place. 
	
	\item We for the first time propose to conduct VAD through building and completing VCTs, which offers a promising alternative to frequently-used reconstruction or frame prediction paradigm.
	
	\item We propose to equip VCTs with the VCT type ensemble and modality ensemble strategy respectively, so as to fully exploit the temporal context and motion information in video events.
	
	\item We further propose localization-level, event-level, model-level and decision-level solutions respectively to further enhance VCC, which enables us to fully exploit VCC's potential and obtain evident performance gain.
    	
	
	
\end{itemize}
Extensive empirical evaluations have demonstrated that our approach achieves evidently superior performance to state-of-the-art VAD methods. The rest of our paper is organized as follows: Sec. \ref{sec:related_work} will review classic VAD methods and DNN based VAD methods respectively; Sec. \ref{sec:method} presents the framework of basic VCC in detail; Sec. \ref{sec:evcc} introduces practical strategies to improve the basic VCC into the enhanced VCC; Sec. \ref{sec:eval} presents the results of empirical evaluations on various benchmark datasets and discuss each component of the proposed approach; Sec. \ref{sec:conclusion} concludes this paper.

\section{Related Work}\label{sec:related_work}

\subsection{Classic VAD}
\label{sec:classic_review}
Classic VAD methods are usually comprised of two stages: The feature extraction stage based on carefully designed handcrafted feature descriptors, as well as a separated VAD stage based on classic anomaly detection methods. As for the feature extraction stage, early methods rely on traditional object detection \cite{Viola2001Robust} or tracking \cite{Yilmaz2006Object} technique to extract high-level features of foreground objects, such as motion trajectory \cite{zhang2009learning, piciarelli2008trajectory} and destination \cite{basharat2008learning}. However, those methods are prone to fail in crowded scenes \cite{mahadevan2010anomaly}, which severely limits their application to many real-world scenarios. To handle the crowded case, low-level features that describe video events by pixel-level statistics have been thoroughly explored for feature extraction, \textit{e.g.} dynamic texture \cite{mahadevan2010anomaly}, histogram of optical flow \cite{cong2011sparse}, spatio-temporal gradients \cite{lu2013abnormal, kratz2009anomaly}, 3D SIFT \cite{cheng2015video}, etc. Despite much effort for feature descriptor design, it is still the major performance bottleneck for classic VAD methods: Manually designed descriptors often suffer from limited discriminative power and poor transferability among different scenes, not to mention that feature engineering requires domain expertise as well as massive time and labour. In the subsequent VAD stage, features of video events are fed into a classic anomaly detection method to model normality and discern anomalies. Various methods have been explored for this purpose, such as probabilistic models \cite{mahadevan2010anomaly,cheng2015video}, sparse coding and its variants \cite{cong2011sparse,lu2013abnormal,zhao2011online}, one-class classifier \cite{yin2008sensor}, sociology or nature inspired models \cite{Mehran2009Abnormal,2016Online}, etc. Since two stages in classic VAD are decoupled, there is also no guarantee on the compatibility of the anomaly detection method and the extracted features, which could be suboptimal for VAD.

\begin{figure*}
	\centering
	\includegraphics[scale=0.85]{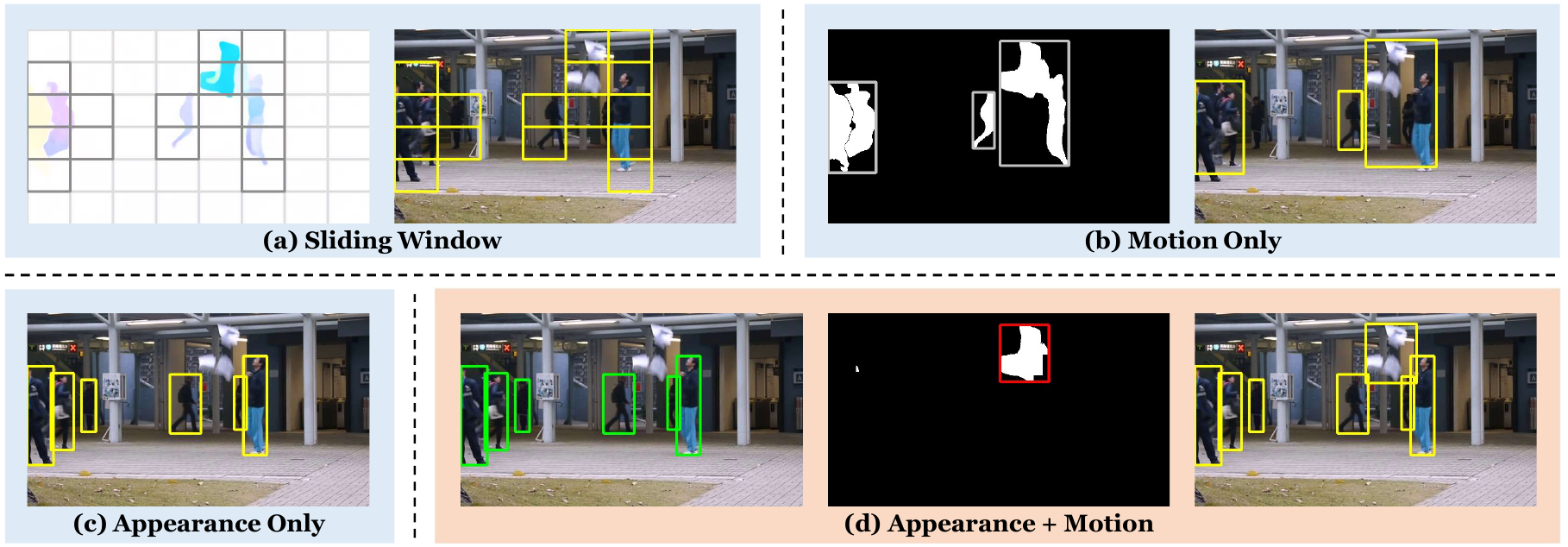}
	\caption{Comparison of RoI localization: Sliding window (a) or motion only (b) produces imprecise localization, while appearance only (c) yields non-comprehensive localization results. The proposed pipeline aims to achieve precise and comprehensive localization simultaneously (d).}
	\label{fig:visual_vee}
\end{figure*}

\subsection{DNN Based VAD}
\label{sec:dnn_review}
Instead of extracting features from video events by manually designed descriptors, DNN based VAD aims to learn proper features automatically from video events via DNNs. Learned features can be either fed into a classic anomaly detection method, or directly used for end-to-end VAD. With only roughly labeled normal videos for training, most DNN based VAD methods follow a \textit{reconstruction}  or \textit{frame prediction} paradigm: \textbf{(1)} Reconstruction based methods learn to reconstruct normal video events in training, and assume that a poor reconstruction indicates the emergence of an abnormal event. Deep autoencoder (AE) and its variants are the most frequently-used model for reconstruction: The pioneer work from Xu \textit{et al.} \cite{xu2015learning} introduces fully-connected stacked denoising AE to address VAD, and its improved version is reported in \cite{xu2017detecting}; Hasan \textit{et al.} \cite{hasan2016learning} leverage convolutional AE (CAE) as an alternative to AE, since CAE is more suitable for modeling images and videos. Then, numerous CAE variants are explored in recent research, such as Winner-take-all CAE \cite{tran2017anomaly} and Long Short Term Memory based CAE \cite{luo2017remembering}; Yan \textit{et al.} \cite{yan2018abnormal} explore the marriage of variational AE (VAE) and VAD, which is achieved by integrating VAE into a two-stream recurrent framework. Wang \textit{et al.} \cite{Wang2019GenerativeNN} propose to combine VAE and UNet to achieve more accurate pixel-wise reconstruction; Abati \textit{et al.} \cite{abati2019latent} propose a combination of AE and a parametric density estimator; Gong \textit{et al.} \cite{Gong_2019_ICCV} devise a memory-augmented AE to enable the reconstruction to be more discriminative. Besides, the cross-modality reconstruction that aims to learn the appearance-motion correspondence \cite{nguyen2019anomaly} is shown to be promising. Apart from AE, other types of DNNs like sparse coding based recurrent neural network (RNN) \cite{luo2017revisit,zhou2019anomalynet,Luo2021VideoAD} and generative adversarial network \cite{ravanbakhsh2017abnormal,sabokrou2018adversarially,Zaheer2020OldIG} are also explored for reconstruction based VAD. Recently, other techniques are utilized to incorporate AE to achieve reconstruction based VAD, such as fast sparse coding \cite{Wu2020FastSC}, deep embedded clustering \cite{Markovitz2020GraphEP}, deep k-means \cite{Chang2020ClusteringDD}, deep support vector domain description \cite{Wu2020ADO}. 
\textbf{(2)} \textit{Frame prediction} based methods learn to predict current frames by previous frames, while a poorly predicted frame is assumed to contain anomalies. Liu \textit{et al.} \cite{liu2018future} for the first time validate frame prediction as a useful baseline for DNN based VAD, and they also impose appearance and motion constraints to guarantee the quality of predicting normal events. Afterwards, Lu \textit{et al.} \cite{lu2019future} improve prediction by a convolutional variational RNN model. Since prediction on a per-frame basis leads to the bias towards background \cite{liu2019exploring}, Zhou \textit{et al.} \cite{zhou2019attention} introduce the attention mechanism in prediction. Other methods are also proposed to enhance prediction, such as bidirectional prediction \cite{Fang2020AnomalyDW}, multi-timescale prediction \cite{Rodrigues2020MultitimescaleTP}, multi-space prediction \cite{Zhang2020NormalityLI}, multi-path prediction \cite{Wang2020RobustUV}.
Another natural instinct is to combine prediction with reconstruction into a hybrid paradigm: Zhao \textit{et al.} \cite{zhao2017spatio} design a spatio-temporal CAE that consists of an encoder and two decoders, which conduct reconstruction and prediction respectively; Morais \textit{et al.} \cite{morais2019learning} propose to simultaneously reconstruct and predict human skeletons by a message-passing encoder-decoder RNN; Ye \textit{et al.} \cite{ye2019anopcn} incorporate reconstruction into prediction by a predictive coding network based framework; Park \textit{et al.} \cite{Park2020LearningMN} propose a new memory module to record normal prototypical patterns, which is deployed in the reconstruction and prediction networks respectively.
In addition to reconstruction and frame prediction, other DNN based methods are also explored. Hinami \textit{et al.} \cite{hinami2017joint} propose to detect and recount abnormal events by integrating a generic model and environment-dependent anomaly detectors. Wang \textit{et al.} \cite{Wang2020ClusterAC} introduce contrastive learning to VAD. Other methods \cite{Pang2020SelfTrainedDO} typically extract features from a pre-trained model then feed them into classic techniques such as iterative quantization hashing \cite{Ravanbakhsh2018PlugandPlayCF} and one-class support vector machine \cite{Ionescu2019DetectingAE}.

\section{Basic VCC}
\label{sec:method}

In this section, we will introduce the basic framework of the proposed VAD paradigm: Visual cloze completion (VCC). Basic VCC is made up of two essential components: Video event extraction and visual cloze tests (VCTs). The motivation and details of each component will be presented below.   

\begin{figure*}[ht]
	\centering
	\includegraphics[scale=0.95]{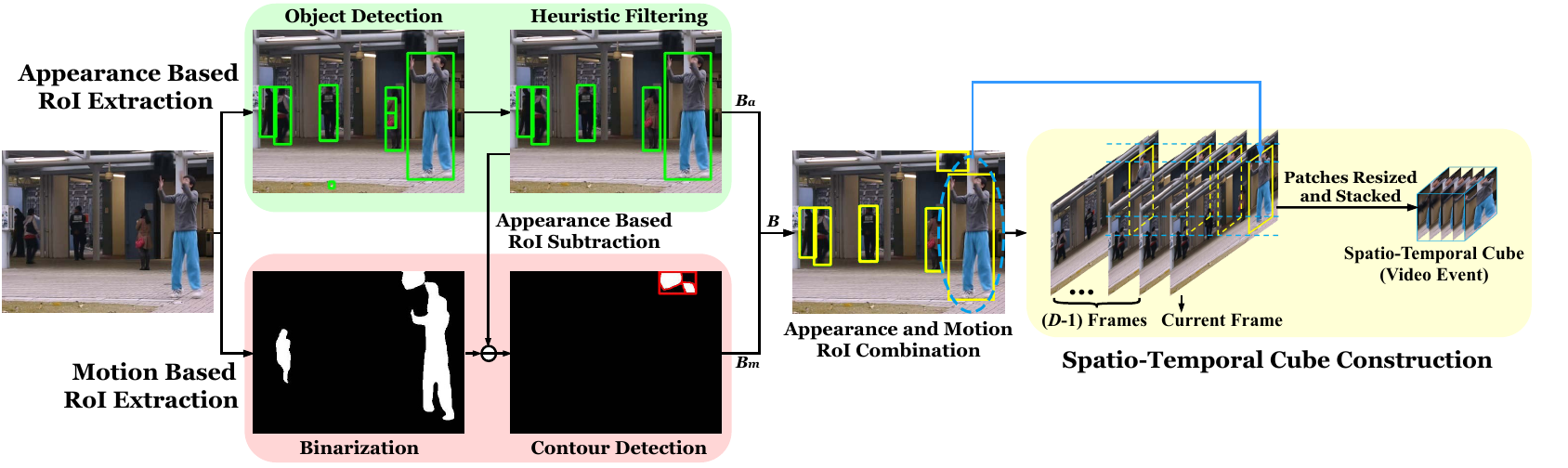}
	\caption{Pipeline of video event extraction: \textbf{(1)} Appearance based RoI extraction (green): Appearance based RoIs are extracted with a pre-trained object detector and filtered based on efficient thresholding. \textbf{(2)} Motion based RoI extraction (red): First, motion map is binarized by magnitude into a binary map. Then, highlighted pixels in appearance based RoIs are subtracted from the binary map. Finally, contour detection and thresholding are applied to the binary map to obtain motion based RoIs. \textbf{(3)} Spatio-temporal cube (STC) extraction (yellow): For each RoI, patches from current frame and $(D-1)$ previous frames are extracted. $D$ patches are then resized and stacked into a STC, which encloses a video event.}
	\label{fig:event_extraction}
\end{figure*}

\subsection{Video Event Extraction}
\subsubsection{Overview}
\label{sec:vee_overview}
An appropriate representation of video events is the foundation for good VAD performance. To this end, we simply assume that a video event is supposed to include a subject (\textit{i.e.} a foreground object) and its activity in a temporal interval. Therefore, a natural solution is to enclose a video event by a spatio-temporal cube (STC) denoted by $V$. To build a STC, the spatial region of the subject on the video frame, which is viewed as the region of interest (RoI) here, should be marked by a bounding box. With the location $b$ of this RoI, a patch sequence $(p_1,\cdots, p_D)$ with $D$ patches is extracted from the current and $(D-1)$ temporally adjacent frames to describe the activity of this subject. Since DNNs usually takes fixed-size inputs, we resize those patches into $h\times w$ new patches $(p'_1, \cdots, p'_D)$ and stack them into a $h\times w\times D$ STC: $V=[p'_1; \cdots; p'_D]$. In this paper, $D$ is often set to a small value like $5$ or $10$ to represent a small interval, which facilitates us to assume that the subject of video event safely stays in the RoI during the temporal interval.

\subsubsection{Motivation}

To extract high-quality STCs to represent video events, the key is to localize RoIs of foreground objects, which then makes it possible to extract corresponding video events. In this paper, we argue that the localization should be both {\textit{precise}} and {\textit{comprehensive}}. To be more specific, the \textit{precise} localization expects the whole region of a foreground object to be covered by a compact bounding box, while the bounding box contains minimal irrelevant background; The \textit{comprehensive} localization requires all foreground objects to be extracted without omission. However, as we explained in the first issue of Sec. \ref{sec:intro}, it is hard for existing VAD methods to realize precise and comprehensive localization simultaneously, and we intuitively illustrate this in Fig. \ref{fig:visual_vee}: The classic sliding window strategy often splits one foreground object by several windows (Fig. \ref{fig:visual_vee} (a)); Motion based localization cannot discriminate different objects and extract excessive irrelevant background (Fig. \ref{fig:visual_vee} (b)); The object detector that only uses appearance cues tends to omit novel or blurring objects (Fig. \ref{fig:visual_vee} (c)). Thus, such localization prevents DNNs from building a good normal event model and undermines VAD performance. In other words, a both precise and comprehensive RoI localization (Fig. \ref{fig:visual_vee} (d)) is an inevitable prerequisite for good VAD performance.

To this end, we recall that a video event is defined to be a foreground object and its activity. Thus, both \textit{appearance cues} from objects and \textit{motion cues} from their activities need to be considered for extracting RoIs. As to appearance cues, the impressive success of modern object detection \cite{2020Deep} naturally motivates us to leverage a generic object detector, which can exploit appearance cues efficiently for localization. With generic knowledge from large-scale real-world datasets like Microsoft COCO \cite{2014Microsoft}, the pre-trained detector is able to extract the majority of daily objects (\textit{e.g.} humans and vehicles) in a highly precise manner. However, as illustrated in Sec. \ref{sec:intro}, RoI extraction with only appearance cues is non-comprehensive due to the fatal ``closed-world'' problem. To this end, motion cues provide valuable complementary information to localize omitted foreground objects, which enables us to overcome the ``closed-world'' problem and accomplish more {comprehensive} RoI extraction. More importantly, it needs to be noted that motion based RoI extraction should not be an isolated process-- RoIs that are already localized by appearance cues should be filtered when exploiting motion cues, which can reduce redundant computation and encourage the localization of omitted foreground objects to be more precise. Inspired by those ideas, we propose a new pipeline with both appearance based and motion based RoI extraction, which is shown in Fig. \ref{fig:event_extraction}. 

\begin{algorithm}[t]
    \begin{footnotesize}
	\caption{Proposed RoI extraction pipeline}
	\label{extract_roi}
	\begin{algorithmic}[1] 
		\State {\bfseries Input:} Frame $I_a$ and its motion map $I_m$, pre-trained object detector $M$, threshold $T_s, T_a, T_o, T_b, T_{ar}$
		\State {\bfseries Output}: RoIs represented by a bounding box set $B$
		\State $B_{ap}\gets ObjDet(I_a, M, T_s)$ \quad \# Object detection
		\State $B_{a}=\{\}$ \quad \# Rule based filtering
		\For {$b_{ap} \in B_{ap}$}
		\If{$Area(b_{ap})>T_{a}$ \& $Overlap(b_{ap}, B_{ap})<T_o$}
		\State $B_{a}=B_{a} \cup \{b_{ap}\}$
		\EndIf
		\EndFor
		\State $I^{(b)}_m\gets Bin(I_m, T_b)$ \quad \# Motion map binarization
		\State $I^{(b)}_m\gets RoISub(I^{(b)}_m, B_a)$ \quad \# Remove RoIs in $B_a$
		\State $\mathcal{C}\gets ContourDet(I^{(b)}_m)$ \quad \# Contour detection
		\State $B_{m}=\{\}$
		\For {$c \in \mathcal{C}$}
		\State $b_{m}=ContourBox(c)$ \quad \# Get contour bounding box
		\mathchardef\mhyphen="2D
		\If {$Area(b_{m})>T_{a} \& \frac{1}{T_{ar}}<AspectRatio(b_{m})<T_{ar}$}
		\State $B_{m}=B_{m} \cup \{b_{m}\}$
		\EndIf
		\EndFor
		\State $B=B_{a} \cup B_{m}$
	\end{algorithmic}
	\end{footnotesize}
\end{algorithm}

\subsubsection{Appearance Based RoI Extraction} 

Given a raw video frame $I_a$ and a pre-trained object detector model $M$, the goal of appearance based RoI extraction is to obtain a RoI set $B_a$ via appearance cues of the foreground objects, where $B_a\subseteq \mathbb{R}^4$ and each entry $b_{ap}\in B_a$ refers to a RoI marked by a bounding box. Note that a bounding box is denoted by the coordinates of its top-left and bottom-right vertex, which is a 4-dimensional vector. As shown by the green module in Fig. \ref{fig:event_extraction}, we first feed $I_a$ into $M$, and obtain a preliminary RoI set $B_{ap}$ by selecting those output bounding boxes with confidence scores above the threshold $T_s$. The output class labels from $M$ are discarded, \textit{i.e.} $M$ is only used to provide localization and no fine-grained class information is exploited. Then, we introduce two efficient rules to filter RoIs that are evidently unreasonable: \textbf{(1)} RoI area threshold $T_a$ that filters out overly small RoIs. \textbf{(2)} Overlapping ratio $T_o$ that removes RoIs that are nested or significantly overlapped with larger RoIs in $B_{ap}$. In this way, we ensure that extracted RoIs by appearance cues can precisely localize most foreground objects of daily events.

\subsubsection{Motion Based RoI Extraction} 
\label{sec:motion_roi}
To localize those foreground objects outside the ``closed world'', motion based RoI extraction aims to yield a supplementary bounding box set $B_m$ based on motion cues. Specifically, as shown by the red module of Fig. \ref{fig:event_extraction}, we first introduce a motion map $I_m$, which contains the motion magnitude of each pixel on the current frame, as our motion cues. To obtain $I_m$, the most straightforward way is to compute temporal gradients with two consecutive frames, and other sophisticated means can also be explored for better computation of $I_m$ (discussed in Sec. \ref{sec:of_cue}). With such a motion map, we can simply binarize the motion map by a threshold $T_b$ and yield a binary map that indicates RoI regions with intense motion. Instead of directly applying the binary map, we propose to subtract appearance based RoIs $B_a$ from the map, which benefits motion based RoI extraction in two ways: First, the subtraction of appearance based RoIs enables us to focus on those omitted foreground objects and produce more precise RoIs for them, otherwise the overlap of multiple objects will jointly produce large and imprecise RoIs (see Fig. \ref{fig:visual_vee} (b)). Second, the subtraction can avoid redundant computation. Finally, we propose to perform contour detection to yield the contour and its corresponding bounding box $b_m$, while simple filtering rules (RoI area threshold $T_a$ and maximum aspect-ratio threshold $T_{ar}$) are used to obtain the RoI set $B_m$. Based on two complementary RoI sets, the final RoI set is yielded by $B=B_a\cup B_m$, and the entire RoI extraction process is summarized by Algorithm \ref{extract_roi} and Fig. \ref{fig:event_extraction}. With those RoIs, we are able to extract high-quality STCs to represent video events, and then construct VCTs for DNNs to solve. 

\subsection{Visual Cloze Tests (VCTs)}
\label{sec:vct}
\subsubsection{Motivation}
Having extracted video events in a precise and comprehensive manner, the next step is to learn a DNN based normality model with normal video events. However, just as we discussed by the second issue in Sec. \ref{sec:intro}, commonly-used reconstruction or frame prediction paradigm cannot fully exploit video semantics and temporal context information. To remedy this problem, we propose a new learning paradigm that trains DNNs to complete \textit{visual cloze tests (VCTs)}. VCTs play a central role in our VCC approach. 

\begin{figure*}[ht]
	\centering
	\includegraphics[scale=0.97]{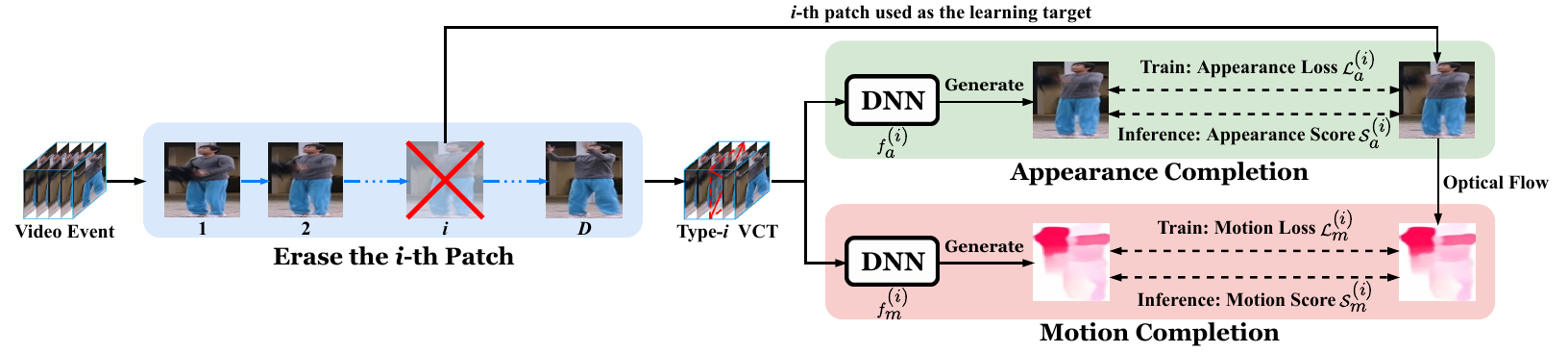}
	\caption{ The basic procedure of VCC: \textbf{(1)} Constructing a type-$i$ VCT (blue): The $i$-th patch of a STC is erased to build a type-$i$ VCT, while the erased patch is used as the target of appearance completion. \textbf{(2)} Appearance completion (green): To complete the VCT, a DNN takes the patches in VCT as input and learns to generate the erased patch. \textbf{(3)} Motion completion (red): A DNN takes the VCT as input and learns to generate the optical flow patch that corresponds to the erased patch.}
	\label{fig:event_completion}
\end{figure*}

VCTs are inspired by the \textit{cloze test}, an extensively-used exercise in language education. It requires students to complete an incomplete text, in which certain words or phases are deliberately erased. In this way, cloze test can test students' grasp of the semantics in words or phrases, as well as their ability to exploit the information of context \cite{wiki_cloze_test}. In the realm of natural language processing (NLP), a similar idea has been explored as an effective pre-training technique, so as to enable DNNs to learn richer semantics from texts \cite{devlin2019bert}. Considering that video semantics and context information are also of paramount importance to discriminating abnormal video events, we are naturally inspired to design VCTs as a counterpart of cloze test in computer vision. Since we assume a video event to be enclosed by a STC, the patch sequence of the STC naturally corresponds to a visual ``sentence'' that describes the video event, while a patch $p'_i$ can be viewed a visual ``word''. With such an analog, a VCT can be built by erasing any patch $p'_i$ from a STC. To complete the VCT, DNNs are required to give a inferred patch $\tilde{p}'_i$, which is supposed to be as close to $p'_i$ as possible. Such a learning paradigm benefits VAD in two aspects: \textbf{(1)} In order to complete such a VCT, DNNs are encouraged to capture video semantics in STC. For example, consider a video event that describes a walking person. DNNs must attend to the motion of some key high-level parts (e.g. the forwarding leg and swinging arm) in STC to realize a good completion. This makes VCT a more meaningful task than reconstruction, as the latter tends to memorize low-level details for training loss reduction. \textbf{(2)} Since STC is the basic processing unit in VCC and any patch in a STC can be erased to create a VCT, we can readily build multiple VCTs by erasing every possible patch. In this way, the temporal context is fully exploited by considering each patch in this context for completion. By contrast, frame prediction based VAD methods only consider prediction errors of a single frame to evaluate anomalies, which involves limited temporal context information in video events. Generally speaking, completing a VCT essentially requires to solve two tasks, \textit{appearance completion} and \textit{motion completion}, which are then equipped with two ensemble strategies. With the illustration in Fig. \ref{fig:event_completion}, we will detail each facet of VCTs below. 

\subsubsection{Appearance Completion}
Given the $j$-th video event represented by the STC $V_j=[p'_{j,1}; \cdots; p'_{j,D}]$, we first erase the $i$-th patch $p'_{j,i}$ of $V_j$ to build a VCT $V_j^{(i)}=[p'_{j,1};\cdots p'_{j,i-1}; p'_{j,i+1};\cdots p'_{j,D}]$, $i\in \{1,\cdots D\}$ (blue module in Fig. \ref{fig:event_completion}). It should be noted that any VCT built by erasing the $i$-th patch of a STC is called a \textit{type-$i$ VCT}, and all {type-$i$ VCTs} are collected as the \textit{type-$i$ VCT set} $\mathcal{V}^{(i)}=\{V_1^{(i)}, \cdots V_N^{(i)}\}$, where $N$ is the number of extracted video events (STCs). Afterwards, as shown by red module in Fig. \ref{fig:event_completion}, a type-$i$ VCT $V_j^{(i)}$ in $\mathcal{V}^{(i)}$ and its corresponding erased patch ${p}'_{j,i}$ are used as the input and learning target respectively to train a generative DNN $f^{(i)}_a$, which aims to generate a patch $\Tilde{p}'_{j,i}=f^{(i)}_a(V_j^{(i)})$ to fill the blank of VCT $V_j^{(i)}$. $f^{(i)}_a$ can be implemented by multiple network architectures, \textit{e.g.} a standard UNet used in basic VCC, and we will explore a more sophisticated solution later (see Sec. \ref{sec:clstm_unet}). To train $f^{(i)}_a$, we minimize the \textit{appearance loss}, which computes the difference between the erased patches and inferred patches for type-$i$ VCT set $\mathcal{V}^{(i)}$:
\begin{equation}
\label{eq:app_loss}
\mathcal{L}^{(i)}_a = \frac{1}{N}\sum^N_{j=1}\Vert \Tilde{p}'_{j,i}-{p}'_{j,i}\Vert^p_p
\end{equation}

Note here we slightly abuse the notation by defining ${p}'_{j,i}$/$\Tilde{p}'_{j,i}$ to be the column vector yielded by concatenating all columns of the original 2D patch ${p}'_{j,i}$/$\Tilde{p}'_{j,i}$, and $\Vert\cdot\Vert_p$ denotes the $p$-norm of a vector. Since the goal of appearance completion is normalized small patches, we discover that a simple appearance loss like Eq. \ref{eq:app_loss} is sufficient for yielding high-quality completions for VCT. By contrast, many DNN based VAD methods like \cite{liu2018future} are frame-based and require adversarial training to improve the quality of generation, which is unnecessary for our appearance completion. Empirically, we notice that different values of $p$ actually performs similarly, so we simply adopt the most frequently-used $p=2$ case, \textit{i.e.} mean square error (MSE) loss. It should be noted that the DNN $f^{(i)}_a$ only handles VCTs from the type-$i$ VCT set $\mathcal{V}_j^{(i)}$, which enables $f^{(i)}_a$ to be more specialized and easier to train. Otherwise, the VAD performance will be degraded.

Since DNNs are trained to complete VCTs created by normal events, we view those video events that correspond to poorly completed VCTs as abnormal in inference. To this end, we can flexibly select any score metric  $\mathcal{S}^{(i)}_a(\Tilde{p}'_{j,i}, {p}'_{j,i})$, such as mean square error (MSE) or Peak Signal to Noise Ratio (PSNR) \cite{liu2018future}, to measure the quality of completions and compute the anomaly score of patch ${p}'_{j,i}$. In fact, our preliminary work \cite{yu2020cloze} shows that choosing $\mathcal{S}^{(i)}_a(\Tilde{p}'_{j,i}, {p}'_{j,i})$ to be MSE proves to be very effective to score anomalies, but we will show that a mixed score that combines different metrics can boost the VAD performance (see Sec. \ref{sec:metric_post}).

\subsubsection{Motion Completion}
\label{sec:motion_completion}
Since motion is the other important attribute of videos, we also intend to take motion information into account when building and completing VCTs. For this purpose, dense optical flow can be leveraged as a highly accessible and effective representation of per-pixel motion in videos. Concretely, it estimates the motion displacement $(dx, dy)$ of the pixel at the position $(x, y)$ between two consecutive frames with time interval $dt$, which are assumed to satisfy:

\begin{equation}
P(x, y, t) = P(x+dx, y+dy, t+dt)
\end{equation}
where $P(x,y,t)$ denotes the pixel intensity at the position $(x,y)$ for time $t$. Optical flow can be computed by either classic methods or DNN based methods \cite{2015Optical}. For efficiency, we estimate the dense optical flow by a pre-trained FlowNetv2 model \cite{ilg2017flownet}. With estimated optical flow map of each frame, we can obtain optical flow patches $(o_{j,1}, \cdots o_{j,D})$ that correspond to video patches $(p_{j,1}, \cdots p_{j,D})$ in $V_j$, and resize them into $h\times w$ patches $(o'_{j,1}, \cdots o'_{j,D})$. Motion completion requires a DNN $f^{(i)}_m$ to infer the optical flow patch of the erased patch $p'_{j,i}$ by $V_j^{(i)}$, \textit{i.e.} $\Tilde{o}'_{j,i}=f^{(i)}_m(V_j^{(i)})$, so as to make the inferred optical flow $\Tilde{o}'_{j,i}$ to be as close to $o'_{j,i}$ as possible. Similar to appearance completion, $f^{(i)}_m$ is trained with motion loss $\mathcal{L}^{(i)}_m$:

\begin{equation}
\mathcal{L}^{(i)}_m = \frac{1}{N}\sum^N_{j=1}\Vert \Tilde{o}'_{j,i}-{o}'_{j,i}\Vert^p_p
\end{equation}
Likewise, we also adopt $p=2$ for $\mathcal{L}^{(i)}_m$, and use the same way to define the motion anomaly score $\mathcal{S}^{(i)}_m(\Tilde{o}'_{j,i}, {o}'_{j,i})$ during inference. With motion completion, we encourage the DNN to infer the motion statistics from the temporal context provided by VCTs, which enables it to consider richer video semantics like appearance-motion correspondence of foreground objects. The process of both appearance and motion completion for a type-$i$ VCT are shown in Fig. \ref{fig:event_completion}.

\subsubsection{Ensemble Strategies}
Ensemble is a powerful technique that combines multiple models into a stronger one \cite{dietterich2000ensemble}. We propose to equip VCTs with two ensemble strategies, so as to fully unleash its potential: \textbf{(1)} \textit{VCT type ensemble}. To fully exploit the temporal context for VAD, each patch in the temporal context of a video event should be involved when computing the video event's anomaly score. To this end, we notice that one STC will produce $D$ different VCTs, thus making it possible to consider each patch in the temporal context for completion. Therefore, we propose to compute the final appearance anomaly score for a video event by an ensemble of scores, which are obtained by completing all different types of VCTs created from this event:

\begin{equation}
\mathcal{S}_a(V_j) = \frac{1}{D}\sum^D_{i=1}\mathcal{S}^{(i)}_a(\Tilde{p}'_{j,i}, {p}'_{j,i})
\end{equation}
Likewise, VCT type ensemble is also applicable to computing the final motion score $\mathcal{S}_m(V_j)$:
\begin{equation}
\mathcal{S}_m(V_j) = \frac{1}{D}\sum^D_{i=1}\mathcal{S}^{(i)}_m(\Tilde{o}'_{j,i}, {o}'_{j,i})
\end{equation}
\textbf{(2)} \textit{Modality ensemble}. Since appearance and motion are both important clues for VAD, we need to fuse results from appearance completion and motion completion to yield the overall anomaly score. For simplicity, we use a weighted sum of $\mathcal{S}_a(V_j)$ and $\mathcal{S}_m(V_j)$ to compute the overall anomaly score $S(V_j)$ for a video event $V_j$:

\begin{equation}
\mathcal{S}(V_j)=w_a\cdot\frac{\mathcal{S}_a(V_j)-\Bar{\mathcal{S}}_a}{\sigma_a} + w_m\cdot\frac{\mathcal{S}_m(V_j)-\Bar{\mathcal{S}}_m}{\sigma_m}
\end{equation}
where $\Bar{\mathcal{S}}_a, \sigma_a, \Bar{\mathcal{S}}_m, \sigma_m$ denote the means and standard deviations of appearance and motion scores for all normal events in the training set, which are used to normalize appearance and motion scores into the same scale. In addition to this straightforward weighting strategy, other sophisticated late fusion strategies like \cite{xu2015learning,xu2017detecting} are also applicable to achieve better modality ensemble performance. For frame scoring, the maximum of all events' scores on a frame is viewed as the frame score.

\section{Enhanced VCC}
\label{sec:evcc}
Basic VCC has already been able to achieve fairly satisfactory VAD performance, but there is still room for improvement. In this section, we will elaborate the localization-level, event-level, model-level and decision-level solutions respectively to enhance VCC. Those solutions can be combined into an enhanced version of VCC, which can achieve significant performance gain.  

\subsection{Localization-level Enhancement}
\label{sec:of_cue}

\begin{figure}
	\centering
	\includegraphics[scale=0.86]{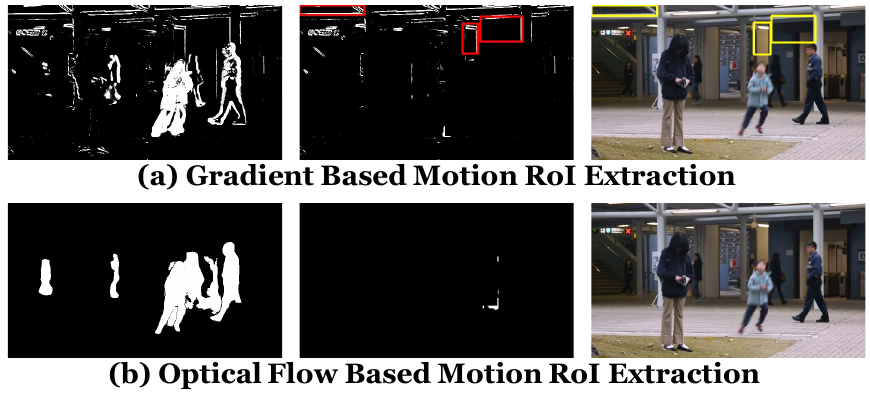}
	\caption{Comparison of temporal gradients and optical flow as motion cues for motion based RoI extraction.}
	\label{fig:GD_OF_comparison}
\end{figure}

As we have introduced in Sec. \ref{sec:motion_roi}, motion based RoI extraction requires to compute the motion map $I_m$, which is yielded by computing temporal gradients in our preliminary work \cite{yu2020cloze}. Nevertheless, computing $I_m$ by temporal gradients suffer from two major drawbacks: First, the appearance of foreground objects will impose a significant influence on the magnitude of temporal gradients, which makes them less reliable to reflect motion. For example, two pedestrians with the same speed may produce different temporal gradients when they are in clothes with different colors. Second, temporal gradients are susceptible to low-level noises. These noises can be pervasive in real-world videos due to various factors like illumination changes or gentle vibration of camera. Disturbed by low-level noises, temporal gradients could generate massive low-level artifacts in motion map (see Fig. \ref{fig:GD_OF_comparison} (a)). Despite the rule based filtering, some artifacts are still misinterpreted as RoIs.

Motivated by observations above, we propose to employ optical flow as more accurate motion cues in this paper. When compared with temporal gradients, optical flow is blessed with several strengths: First, optical flow is less sensitive to appearance, as it is based on correspondence rather than intensity changes. Therefore, it can be a more accurate clue to reflect motion. Second, optical flow, which is estimated by pre-trained FlowNetv2 model, is more robust to low-level noises, owing to FlowNetv2's correlation layer for high-level features and stacked architecture for noise reduction \cite{ilg2017flownet}. Third, optical flow has already been computed as targets for motion completion (see Sec. \ref{sec:motion_completion}), so no additional computation is actually required. By contrast, as shown in Fig. \ref{fig:GD_OF_comparison} (b), we can yield a smoother binary map with less low-level artifacts to indicate motion regions when the optical flow map is used as $I_m$, and misinterpreted RoIs are effectively removed. As a result, improving motion cues by optical flow is able to enhance VCC by more precise motion based RoI localization, which leads to better performance. Besides, optical flow's robustness also makes it easier to determine the binarization threshold $T_b$, as value of $T_b$ can be unified among different datasets regardless of the differences in objects' appearance and scenarios. Note that one can also explore more effective motion cues.

\subsection{Event-level Enhancement}

\begin{figure}[t]
	\centering
	\includegraphics[scale=0.8]{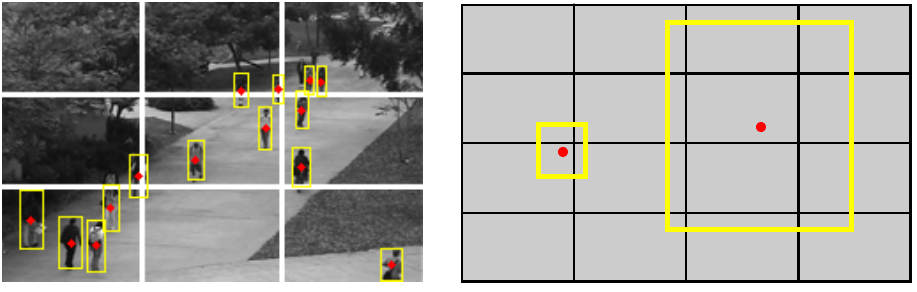}
	\caption{The spatially-localized strategy for block-based learning and inference for enhanced VCC.}
	\label{fig:block}
\end{figure}

\begin{figure*}[ht]
	\centering
	\includegraphics[scale=1.0]{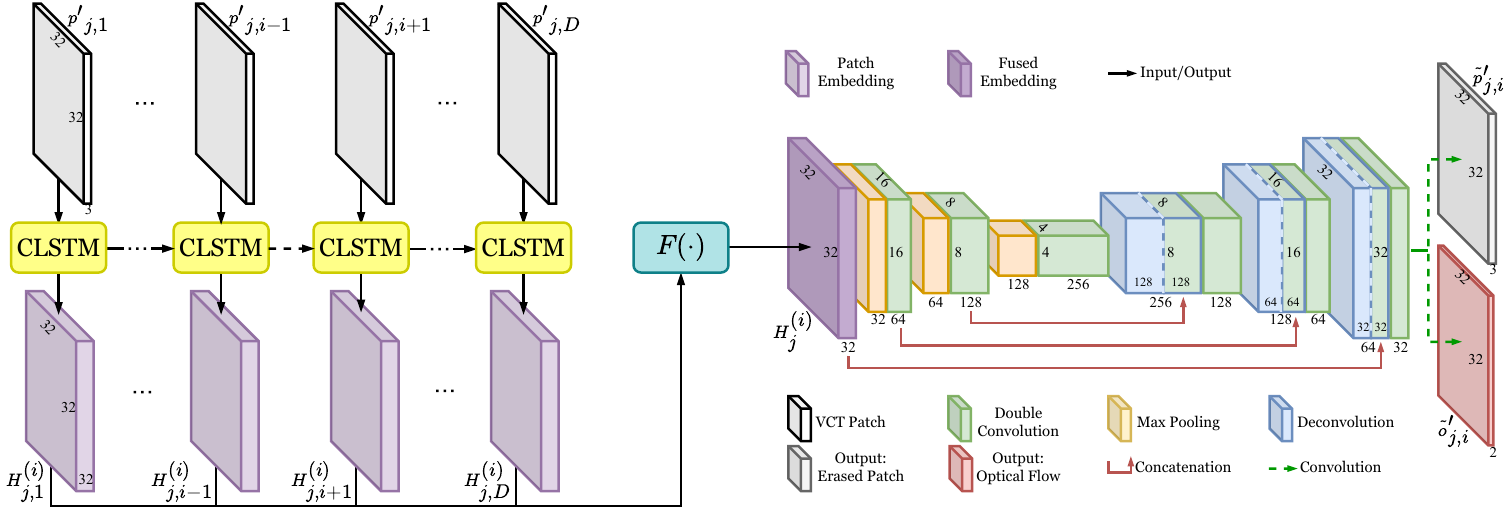}
	\caption{ST-UNet architecture used in our experiments, wherein CLSTM share all parameters. Appearance completion network $f^{(i)}_a$  and motion completion network $f^{(i)}_m$ share the same ST-UNet architecture, except that $f^{(i)}_a$ has 3 output channels corresponding to erased patch $p'_{j,i}$, while $f^{(i)}_m$ has 2 output channels corresponding to optical flow $o'_{j,i}$.}
	\label{fig:ST-UNet}
\end{figure*}

VAD in many real-world scenarios is subject to varied foreground depth. With different depth, the same type of video events may exhibit different sizes and scales, which pose a important challenge to modeling and inference. For example, in a typical scene from UCSDped1 dataset (see Fig. \ref{fig:block}), pedestrians at the bottom left corner have obviously larger size and motion (optical flow) magnitude than those at the top right corner. Hence, video events from the same category (pedestrian walking) seem to possess different characteristics, which enlarges the intra-class difference and undermines the one-class learning for VAD. The proposed video event extraction is able to alleviate this problem to a certain extent by normalizing all RoIs into the same size. However, it does not address this problem from the root for two reasons: First, the normalization is performed spatially, while the motion magnitude cannot be adjusted by spatial interpolation. Second, when up-sampling is performed for normalization, the foreground object will be blurred. 

To mitigate this issue, we design a spatially-localized strategy for training. As shown in Fig. \ref{fig:block}, the core idea of the spatially-localized strategy is to divide the video frame into several local spatial regions, \textit{a.k.a.} blocks. Since each block only cover a local spatial region, we can safely assume that all foreground objects in this block share similar depth. Afterwards, video events in one block are modeled or tested in a separated manner, so as to enable DNNs to only handle video events with a comparable scale. In order to assign a video event to a block, a simple and natural criterion is to assign it to the block that enjoys maximum overlap with the video event's bounding box. Based on this criterion, we propose to introduce a simple theorem below to efficiently determine the assignment of video events:

\begin{theorem}\label{thm:1}
Given an arbitrary $2$-dimensional rectangle $b$ in a $2$-dimensional plane, the plane is uniformly partitioned into rectangular local regions $\{R_i\}^{\infty}_{i=1}$ with any size. If the rectangle $b$'s geometric center $c_b$ and the $k_{th}$ local region $R_k$ satisfy $c_b\in R_k$, the overlap area $\mathcal{O}(b, R_k)$ of $b$ and $R_k$ satisfies:
\begin{equation}
{
 \mathcal{O}(b, R_k)=\sup_i\{\mathcal{O}(b, R_i)\}
}
\end{equation}
The above conclusion can be generalized to any $n$-dimensional hyperspace, where $n$ is an positive integer.
\end{theorem}
The proof of Theorem \ref{thm:1} is given in the Appendix \ref{secA1}. Theorem \ref{thm:1} reveals that we can simply determine if a video event belongs to a block by checking whether the center of its bounding box (red dots in Fig. \ref{fig:block}) lies in this block, since it guarantees a maximum overlap. Note Theorem \ref{thm:1} holds only when the video frame is uniformly partitioned into rectangle blocks like Fig. \ref{fig:block}. However, more fine-grained irregular division is also applicable: One can simply select a single frame from the training videos, and manually divide the frame into several irregular blocks that better describe the depth of different spatial regions, which actually requires minimal cost and labor. Since the surveillance videos usually share fixed background, the division can be fixed in later process. In this paper, we simply adopt regular division for videos with varied foreground depth.

\subsection{Model-level Enhancement}
\label{sec:clstm_unet}

As we discussed in Sec. \ref{sec:vct}, one of our goal is to fully exploit the video semantics and temporal context information for VAD. Motivated by this goal, we propose VCC as a new VAD learning paradigm, which is a paradigm-level solution. However, the standard UNet \cite{ronneberger2015u} used in \cite{yu2020cloze} places more emphasis on the spatial information of a patch, while it does not model the temporal correlation among patches in a STC explicitly. Thus, it is natural to develop a model-level solution that is specifically tailored for this goal. To this end, we design a new DNN architecture named spatio-temporal UNet (ST-UNet), which is more compatiable with our VCC paradigm. Specifically, the core idea of ST-UNet is to synthesize a convolutional long-short-term-memory (CLSTM) \cite{Shi2015ConvolutionalLN} module into the UNet module. For a type-$i$ VCT $V_j^{(i)}=[p'_{j,1};\cdots p'_{j,i-1}; p'_{j,i+1};\cdots p'_{j,D}]$, $i\in \{1,\cdots D\}$, each time the $t_{th}$ patch $p'_{j,t}$ is fed into the CLSTM module to compute $I^{(i)}_{j,t}, F^{(i)}_{j,t}, O^{(i)}_{j,t}$, which correspond the control signals of input gate, forget gate and output gate respectively: 

\begin{equation}
    \begin{aligned}
    I^{(i)}_{j,t} &= sigmoid(W_{pe} \otimes p'_{j,t} + W_{he} \otimes H^{(i)}_{j,t-1} + b_e) \\
    F^{(i)}_{j,t} &= sigmoid(W_{pf} \otimes p'_{j,t} + W_{hf} \otimes H^{(i)}_{j,t-1} + b_f) \\
    O^{(i)}_{j,t} &= sigmoid(W_{po} \otimes p'_{j,t} + W_{ho} \otimes H^{(i)}_{j,t-1} + b_o) 
    \end{aligned}
\end{equation}
where $W_{pe},W_{pf},W_{po},W_{he},W_{hf}, W_{ho}$ are learnable convolutional kernels, and $b_e,b_f,b_o$ denote the associated biases. $H^{(i)}_{j,t-1}$ represents a high-level embedding of the previous patch $p'_{j,t-1}$, which is introduced to involve the influence of temporal history information, while $\otimes$ denotes the convolution operation. With $I^{(i)}_{j,t}, F^{(i)}_{j,t}$ to control the influx and outflux of the past and present information, which are recorded in $C^{(i)}_{t-1}$ and $\Tilde{C}^{(i)}_{j,t}$, the current cell state $C^{(i)}_{j,t}$ in CLSTM module can be calculated as:

\begin{equation}
    \begin{aligned}
    \Tilde{C}^{(i)}_{j,t} &= tanh(W_{pc} \otimes p'_{j,t} + W_{hc} \otimes H^{(i)}_{j,t-1} + b_c) \\
    C^{(i)}_{j,t} &= F^{(i)}_{j,t} \circ C^{(i)}_{t-1} + I^{(i)}_{j,t} \circ \Tilde{C}^{(i)}_{j,t}
    \end{aligned}
\end{equation}
where $W_{pc},W_{hc}$ and $b_c$ denotes the convolutional weights and bias, and $\circ$ is the Hadamard product. With the cell state $C^{(i)}_{j,t}$ and the control signal for output gate $O^{(i)}_{j,t}$, the high-level embedding $H^{(i)}_{j,t}$ of current patch $p'_{j,t}$ is computed by: 

\begin{equation}
    \begin{aligned}
    H^{(i)}_{j,t} &= O^{(i)}_{j,t} \circ tanh(C^{(i)}_{j,t})
    \end{aligned}
\end{equation}
In this way, each patch $p'_{j,t}$ in the VCT $V_j^{(i)}$ is sequentially fed into CLSTM and transformed into a high-level embedding $H^{i}_{j,t}$. $H^{i}_{j,t}$ is not only expected to abstract the current patch into a high-level embedding with richer semantics, but also involve the temporal history information of past patches. Since $H^{i}_{j,t}$ contains richer semantics and temporal context information, we can collect all high-level embeddings and compute an overall embedding $H^{(i)}_j$ for the type-$i$ VCT of $j_{th}$ video event, $V_j^{(i)}$, with a fusion function $F(\cdot)$:

\begin{equation}
H^{(i)}_j = F(H^{(i)}_{j,1},\cdots H^{(i)}_{j,i-1},H^{(i)}_{j,i+1},\cdots H^{(i)}_{j,D})
\end{equation}
$F(\cdot)$ can be implemented by various means, \textit{e.g.} an element-wise operator or a convolution layer, while we simply choose $F(\cdot)$ to be the element-wise summation in this paper. $H^{(i)}_j$ enables us to maximally record the video semantics and temporal context information from the VCT. Therefore, instead of raw patches from STCs,  $H^{(i)}_j$ is then fed into the UNet module to obtain the completion results to the VCT:

\begin{equation}
\begin{aligned}
\Tilde{p}'_{j,i} = U_a(H^{(i)}_j) \\
\Tilde{o}'_{j,i} = U_m(H^{(i)}_j)
\end{aligned}
\end{equation}
where $U_a$ and $U_m$ represents the case for appearance and motion completion respectively. Compared with the preliminary work that adopts standard UNet for VCC \cite{yu2020cloze}, the proposed ST-UNet enjoys three advantages: First, the introduction of CLSTM module enables us to explicitly model the temporal correlation of patches in STCs on the model level; Second, the model can maximally exploit the temporal context by fusing the high-level embedding of all patches in the VCT into an overall embedding; Third, by feeding the overall embedding rather than raw patches into the UNet module, we can encourage DNNs to achieve a better utilization of high-level video semantics for VCT completion. Our later evaluation suggests that using ST-UNet as DNN architecture constantly outperforms UNet in VCC.

\subsection{Decision-level Enhancement}
\label{sec:metric_post}
At the stage of decision, the computation of anomaly score and post-processing could also exert a significant influence on VAD performance. For anomaly score metric, it is shown by our preliminary works \cite{yu2020cloze} that MSE could be an effective anomaly score metric of VCC. However, research has revealed that MSE suffers from some weaknesses, such as its strong emphasis on per-pixel error and negligence of high-level structure. Due to those weaknesses, MSE may overly focus on low-level differences and cannot comprehensively reflect some high-level structural differences between two patches. Therefore, we propose to introduce Structural Similarity (SSIM) as a supplementary score metric to MSE. SSIM is computed as follows:
\begin{equation}
\begin{split}
     & SSIM(p'_{j,i}, \Tilde{p}'_{j,i})  = \\ &\frac{(2\mu_{p'_{j,i}}\mu_{\Tilde{p}'_{j,i}} + c_1)(2\sigma_{p'_{j,i}\Tilde{p}'_{j,i}} + c_2)}{(\mu^2_{p'_{j,i}} + \mu^2_{\Tilde{p}'_{j,i}} + c_1)(\sigma^2_{p'_{j,i}} + \sigma^2_{\Tilde{p}'_{j,i}} + c_2)}
\end{split}
\end{equation}
where $(\mu_{p'_{j,i}}, \sigma_{p'_{j,i}})$ and $(\mu_{\Tilde{p}'_{j,i}},\sigma_{\Tilde{p}'_{j,i}})$ denote the mean and standard deviation of pixel intensity for the erased patch $p'_{j,i}$ and inferred patch $\Tilde{p}'_{j,i}$ respectively. $\sigma_{p'_{j,i}\Tilde{p}'_{j,i}}$ is the covariance between the pixels in patches $p'_{j,i}$ and $\Tilde{p}'_{j,i}$, while $c_1$ and $c_2$ are constant. The final score is computed by a combination of MSE and SSIM, and it is applicable to both appearance and motion completion.

In addition to a mixed anomaly score metric that can better reflect the completion quality, post-processing based score rectification is another effective way to refine the obtained anomaly scores. The motivation for score rectification stems from the observation that video events are continuous, so the anomaly scores of adjacent video frames are supposed to be close. Therefore, it is natural for us to rectify the anomaly score of current video frame with those anomaly scores yielded by previous temporally adjacent frames. Suppose that the anomaly score for the $l_{th}$ frame and its $W$ previous frames are $\mathcal{S}_l, \mathcal{S}_{l-1}, \cdots \mathcal{S}_{l-W}$, we propose the general formulation below to calculate the rectified anomaly score $\hat{\mathcal{S}}_l$ is:

\begin{equation}
     \hat{\mathcal{S}}_l=\frac{1}{Z}\sum^{W}_{m=0}\omega_{m}\mathcal{S}_{l-m}
\end{equation}
where $\omega_m$ is a non-negative weight. $Z=\sum_{m}\omega_m$ is a normalizing factor. There are multiple ways to set the weight $\omega_m$. For example, one can set $\omega_m=q^m$, where $0<q\leq 1$. When $q<1$, the frame that is closer to current frame is considered to be more important for the rectification. When $q=1$, the post-processing is equivalent to the temporal moving average. In addition, one can also set $\omega_m$ to obtain an 1-d Gaussian or median filter. We will compare different types of rectification strategies in later experiments, and the results show that even the simplest form of score rectification can produce well-rectified anomaly scores.

\begin{figure*}[ht]
	\centering
	\includegraphics[scale=0.7]{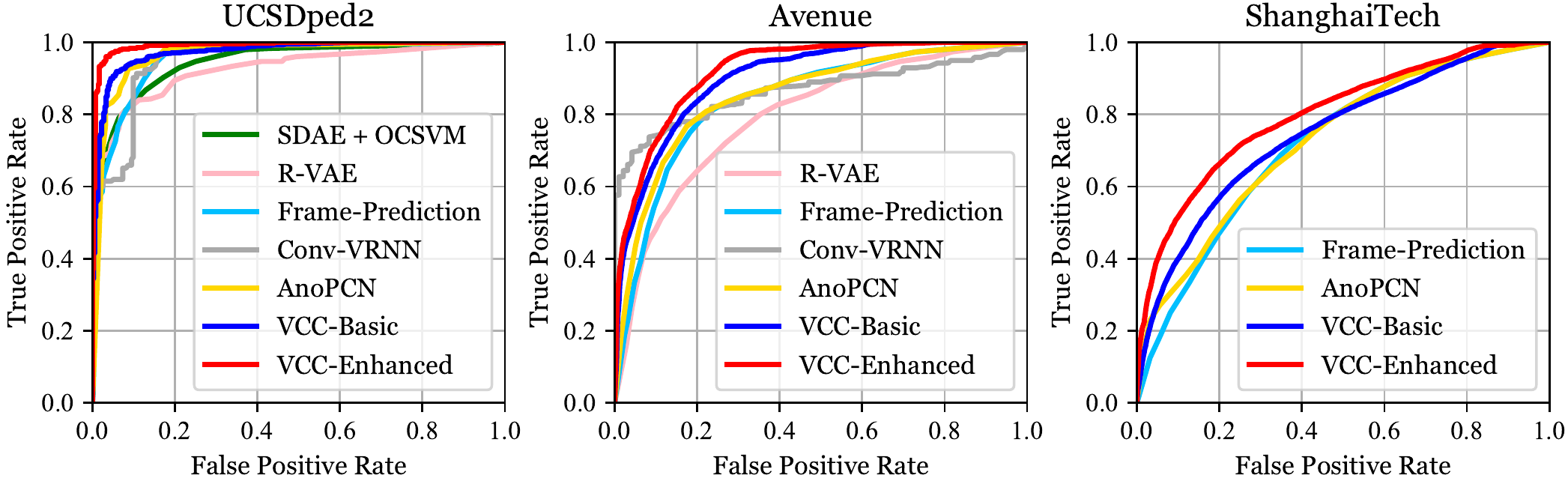}
	\caption{Comparison of frame-level ROC curves.}
	\label{fig:ROC}
\end{figure*}

\begin{table}
	\centering
	\caption{VAD benchmark datasets (``\# of Frames'' is the total number of video frames, ``Resolution'' is the video frame resolution, ``Test split'' indicates whether a separated test set is provided, ``Label'' indicates frame-level or pixel-level labels are provided in the dataset).}
	\linespread{1}\selectfont
	\scalebox{0.8}{
		\begin{tabular}{ccccc}
			\toprule
			 Dataset & \# of Frames & Resolution & Test split & Label   \\
			\midrule
			UCSDped2 \cite{mahadevan2010anomaly} &  4560  &  $240\times 320$ & Yes & pixel  \\
			Avenue \cite{lu2013abnormal} &  30652  &  $360\times 640$ & Yes & pixel  \\
			ShanghaiTech \cite{liu2018future} &  315306  &  $480\times 856$ & Yes & pixel  \\
			UCSDped1 \cite{mahadevan2010anomaly} &  14000  &  $158\times 238$ & Yes & pixel  \\
			UMN \cite{Adam2008Robust} &  7739  &  $240\times 320$ & No &  frame  \\
			Subway Exit \cite{Adam2008Robust} &  64902  &  $384\times 512$ & No & frame  \\
			\bottomrule
	    \end{tabular}}
	\label{tab:benchmark}
\end{table}

\begin{table*}
    \newcommand{\tabincell}[2]{\begin{tabular}{@{}#1@{}}#2\end{tabular}}
	\centering
	\caption{Performance comparison on UCSDped2, Avenue and ShanghaiTech.}
	\linespread{1}\selectfont
	\scalebox{0.75}{
		\begin{tabular}{cc|cccc|cc|cc}
			\toprule
			\multirow{2}{*}{Type} & \multirow{2}{*}{Method} & \multicolumn{4}{c|}{UCSDped2} & \multicolumn{2}{c|}{Avenue} & \multicolumn{2}{c}{ShanghaiTech}\\
			& & AUC(F) & EER(F) & AUC(P) & EER(P) & AUC(F) & EER(F) & AUC(F) & EER(F) \\
			\midrule
			\multirow{7}{*}{\tabincell{c}{Classic Methods}} & SF \cite{Mehran2009Abnormal}  & 55.6\%  & -  & -  & - & - & - & - & - \\
			& MDT \cite{mahadevan2010anomaly}  & 82.9\%  & 25\%  & -  & - & - & - & - & - \\
			 & SR \cite{cong2011sparse}  & 82.9\%  & 25\%  & -  & - & - & - & - & - \\
			 & LSA \cite{2012Video}  & -  & 18\%  & -  & - & - & - & - & - \\
			 & MAC \cite{2016Combining}  & 90\%  & -  & 73.7\%  & - & - & - & - & - \\
			 & GAS \cite{2016Online}  & 94.1\%  & -  & -  & - & - & - & - & - \\
			 & Unmasking \cite{tudor2017unmasking}  & 82.2\%  & -  & -  & - & 80.6\% & - & - & - \\
			\hline
			\multirow{18}{*}{\tabincell{c}{Reconstruction\\Based Methods}} & CAE \cite{hasan2016learning}  & 90.0\% & 21.7\%  & - & - & 70.2\% & 25.1\% & - & - \\
			& AMDN \cite{xu2017detecting} & 90.8\% & 17.0\% & - & - & - & - & - & - \\
			& SRNN \cite{luo2017revisit} & 92.2\% & - & - & - & 81.7\% & - & 68.0\% & - \\
			& WTA-CAE \cite{tran2017anomaly} & 96.6\% & 8.9\% & 89.3\% & 16.9\% & - & - & - & - \\
			& AM-GAN \cite{ravanbakhsh2017abnormal} & 93.5\% & 14.0\% & - & - & - & - & - & - \\
			& ConvLSTM-AE \cite{luo2017remembering} & 88.1\% & - & - & - & 77.0\% & - & - & - \\
			& R-VAE \cite{yan2018abnormal} & 92.4\% & 15.2\% & - & - & 79.6\% & 27.5\% & - & - \\
			& ADV-OC \cite{sabokrou2018adversarially} & - & 13.0\% & - & - & - & - & - & - \\
			& PDE-AE \cite{abati2019latent} & 95.4\% & - & - & - & - & - & 72.5\% & - \\
			& Mem-AE \cite{Gong_2019_ICCV} & 94.1\% & - & - & - & 83.3\% & - & 71.2\% & - \\
			& AM-Corr \cite{nguyen2019anomaly} & 96.2\% & - & - & - & 86.9\% & - & - & - \\
			& AnomalyNet \cite{zhou2019anomalynet} & 94.9\% & 10.3\% & 52.8\% & - & 86.1\% & 22.0\% & - & - \\
			& SRNN-AE \cite{Luo2021VideoAD} & 92.2\% & - & - & - & 83.5\% & - & 69.6\% & - \\
			& GEPC \cite{Markovitz2020GraphEP} & - & - & - & - & - & - & 76.1\% & - \\
			& OGNet \cite{Zaheer2020OldIG} & 98.1\% & 7\% & - & - & - & - & - & - \\
			& Clustering-AE \cite{Chang2020ClusteringDD} & 96.5\% & - & - & - & 86.0\% & - & 73.3\% & - \\
			& FSCN \cite{Wu2020FastSC} & 92.8\% & 12.5\% & - & - & 85.5\% & 20.7\% & - & - \\
			& DeepOC \cite{Wu2020ADO} & 96.9\% & 8.8\% & 95.0\% & - & 86.6\% & 18.5\% & - & - \\
			\hline
			\multirow{10}{*}{\tabincell{c}{Frame Prediction\\Based Methods}} & Frame-Prediction \cite{liu2018future} & 95.4\% & - & - & - & 85.1\% & - & 72.8\% & - \\
			& VRNN \cite{lu2019future} & 96.1\% & - & - & - & 85.8\% & - & - & - \\
			& Att-prediction \cite{zhou2019attention} & 96.0\% & - & - & - & 86.0\% & - & - & - \\ 
			& Multi-Prediction \cite{Rodrigues2020MultitimescaleTP} & - & - & - & - & 82.9\% & - & 76.0\% & - \\
			& SIGNet \cite{Fang2020AnomalyDW} & 96.2\% & - & 48.4\% & - & 86.8\% & - & - & - \\
			& Online \cite{DOSHI2021107865} & 97.2\% & - & - & - & 86.4\% & - & 70.9\% & - \\
			& Multispace \cite{Zhang2020NormalityLI} & 95.4\% & - & - & - & 86.8\% & - & 73.6\% & - \\
			& DMMNet \cite{Li2020VideoFP} & 94.5\% & - & - & - & - & - & - & - \\
			& Bi-Prediction \cite{Chen2020AnomalyDI} & 96.6\% & - & - & - & 87.8\% & - & - & - \\
			& Multipath-Pred. \cite{Wang2020RobustUV} & 96.3\% & - & - & - & 88.3\% & - & 76.6\% & - \\
			\hline
			\multirow{5}{*}{\tabincell{c}{Hybrid of\\Recon. and Pred.}} & ST-CAE \cite{zhao2017spatio} & 91.2\% & 16.7\% & - & - & 80.9\% & 24.4\% & - & - \\
			& MPED-RNN \cite{morais2019learning} & - & - & - & - & - & - & 73.4\% & - \\
			& Mem-Guided \cite{Park2020LearningMN} & 97.0\% & - & - & - & 88.5\% & - & 70.5\% & - \\
			& AnoPCN \cite{ye2019anopcn} & 96.8\% & - & - & - & 86.2\% & - & 73.6\% & - \\
			& Predictive-AE \cite{Lai2020VideoAD} & 95.8\% & - & - & - & 87.4\% & - & - & - \\
			\hline
			\multirow{6}{*}{\tabincell{c}{Other DNN\\Based Methods}} & FCN \cite{2018Deep} & -  & 11\%  & - & 15\% & - & - & - & - \\
			& Recounting \cite{hinami2017joint} & 92.2\% & 13.9\% & 89.1\% & 15.9\% & - & - & - & -\\
			& TCP \cite{Ravanbakhsh2018PlugandPlayCF} & 88.4\% & 18.0\% & - & - & - & - & - & - \\
			& NNC \cite{Ionescu2019DetectingAE} & - & - & - & - & 88.9\% & - & - & - \\
			& Cluster-Att \cite{Wang2020ClusterAC} & - & - & - & - & 87.0\% & - & 79.3\% & - \\
			& Scene-Aware \cite{Sun2020SceneAwareCR} & - & - & - & - & 89.6\% & 21.1\% & 74.7\% & 28.6\% \\
			\hline
			\multirow{2}{*}{\tabincell{c}{Proposed VCC}} & Basic & 97.3\% & 7.5\% & 93.0\% & 12.2\% & 89.6\% & 17.9\% & 74.8\% & 31.5\% \\
			& Enhanced & 99.0\%  & 4.4\%  & 96.8\%  & 6.4\% & 92.2\% & 16.3\% & 80.2\% & 27.0\% \\
			\bottomrule
	    \end{tabular}}
	\label{tab:p2_av_sh}
\end{table*}

\section{Empirical Evaluations}
\label{sec:eval}

\subsection{Experimental Setup}

To evaluate the VAD performance of the proposed VCC approach, we mainly conduct the empirical evaluation on the following VAD benchmark datasets: UCSDped2 \cite{mahadevan2010anomaly}, Avenue \cite{lu2013abnormal} and ShanghaiTech \cite{liu2018future}, which are three most commonly-used benchmark datasets for DNN based VAD. To further validate the effectiveness of VCC, we additionally test our approach on three benchmark datasets: UCSDped1 \cite{mahadevan2010anomaly}, UMN \cite{Adam2008Robust} and Subway Exit \cite{Adam2008Robust}, which are less reported for DNN based VAD but frequently used for evaluating earlier classic VAD methods. The detail information of all benchmark datasets are summarized in Table \ref{tab:benchmark}. The quantitative evaluation of VAD is usually exercised under either the \textit{frame-level criteria} or the \textit{pixel-level criteria} \cite{mahadevan2010anomaly}. To be more specific, frame-level criteria consider a video frame that contains anomalies to be correctly detected if any pixel on the frame is determined to be abnormal. As a comparison, pixel-level criteria are believed to be more accurate by considering the localization of abnormal events \cite{mahadevan2010anomaly}: One abnormal video frame can only be considered to be correctly detected when more than $40\%$ pixels of anomalies on this frame are correctly identified, which avoids the case where the abnormal events are poorly localized. Pixel-level criteria also require pixel-level annotations to be provided. With either of the above criteria, we can compute equal error rate (EER) and area under the curve (AUC) of Receiver Operation Characteristic Curve (ROC) as quantitative performance measure. Following the standard practice in the literature, we perform both frame-level and pixel-level evaluation for UCSDped1 and UCSDped2 dataset, while only frame-level evaluation is performed on other datasets. The implementation details of VCC are reported in Appendix \ref{secA2}, and our codes and results are publicly available for research at {\url{https://github.com/yuguangnudt/VEC_VAD/tree/VCC}}.

\begin{table}[t]
	\centering
	\caption{Performance comparison on UCSDped1.}
	\linespread{1}\selectfont
	\scalebox{0.9}{
		\begin{tabular}{ccccc}
			\toprule
			\multirow{2}{*}{Method} & \multicolumn{2}{c}{Frame-level} & \multicolumn{2}{c}{Pixel-level} \\
			& AUC & EER & AUC & EER  \\
			\midrule
			SF \cite{Mehran2009Abnormal} & 67.5\% & - & 19.7\% & - \\
			MDT \cite{mahadevan2010anomaly} & 81.4\% & 25\% & 44.1\% & 58\% \\
			SR \cite{cong2011sparse} & 89.5\% & 19\% & 50.2\% & 53\% \\
			LSA \cite{2012Video} & 92.7\% & 16\% & - & - \\
			SCL \cite{lu2013abnormal} & 91.8\% & 15\% & 63.8\% & 41\% \\
			GPR \cite{cheng2015video} & 83.8\% & 23.7\% & 63.3\% & 37.3\% \\
			MAC \cite{2016Combining} & 85\% & - & 65\% & - \\
			GAS \cite{2016Online} & 93.8\% & - & 65.1\% & - \\
			Unmasking \cite{tudor2017unmasking} & 68.4\% & - & 52.4\% & - \\
			CAE \cite{hasan2016learning} & 81.0\% & 27.9\% & - & - \\
			WTA-CAE \cite{tran2017anomaly} & 91.9\% & 14.8\% & 68.7\% & 35.7\% \\
			AMDN \cite{xu2017detecting} & 92.1\% & 16.0\% & 67.2\% & 40.1\%  \\
			AM-GAN \cite{ravanbakhsh2017abnormal} & 97.4\% & 8.0\% & 70.3\% & 35.0\% \\
			ConvLSTM-AE \cite{luo2017remembering} & 75.5\% & - & - & - \\
			ST-CAE \cite{zhao2017spatio} & 92.3\% & 15.3\% & - & - \\
			R-VAE \cite{yan2018abnormal} & 75.0\% & 32.4\% & - & - \\ 
			Frame-Prediction \cite{liu2018future} & 83.1\% & - & - & - \\
			TCP \cite{Ravanbakhsh2018PlugandPlayCF} & 95.7\% & 8.0\% & 64.5\% & 40.8\% \\
			AnomalyNet \cite{zhou2019anomalynet} & 83.5\% & 25.2\% & 45.2\%  & - \\
			VRNN \cite{lu2019future} & 86.3\% & - &  &  \\
			Att-prediction \cite{zhou2019attention} & 83.9\% & - & - & - \\
			FSCN \cite{Wu2020FastSC} & 82.4\% & 25.2\% & - & - \\
			DeepOC \cite{Wu2020ADO} & 83.5\% & 23.4\% & 63.1\% & - \\
			SIGNet \cite{Fang2020AnomalyDW} & 86.0\% & - & 51.6\% & - \\ 
			DMMNet \cite{Li2020VideoFP} & 86.0\% & - & - & - \\
			Bi-Prediction \cite{Chen2020AnomalyDI} & 89.0\% & - & - & - \\
			Multipath-Pred. \cite{Wang2020RobustUV} & 83.4\% & - & - & - \\
			\hline
			VCC & 87.7\% & 20.5\% & 76.3\% & 27.9\% \\
			\bottomrule
	    \end{tabular}}
	\label{tab:ped1}
\end{table}

\subsection{Performance Comparison}
\label{sec:comp_stoa}

\subsubsection{Commonly-used Datasets}

To facilitate the comparison with vast VAD methods, our empirical evaluation is mainly carried on three most commonly-used benchmark datasets in DNN based VAD: UCSDped2, Avenue and ShanghaiTech. Within the scope of our best knowledge, we have conducted a comprehensive comparison with all sorts of VAD methods in the literature, including both classic VAD methods and state-of-the-art DNN based VAD methods. DNN based VAD methods can be further categorized into four types:  Reconstruction based methods, frame prediction based methods, hybrid methods that combine reconstruction and prediction, as well as other DNN based methods. Note that we exclude \cite{ionescu2019object} from comparison as it actually uses a different evaluation metric from commonly-used frame-level AUC, which leads to an unfair comparison. As to the proposed VCC, we report the VAD performance of the enhanced VCC in this work, while the performance of basic VCC in our preliminary work \cite{yu2020cloze} is also listed for a reference. The results of comparison are shown in Table \ref{tab:p2_av_sh}, while we also visualize typical frame-level ROC curves in Fig. \ref{fig:ROC} for an intuitive comparison. With those results, we are able to draw the following observations: \textbf{(1)} First, despite that a variety of VAD methods have been developed in literature, the proposed enhanced VCC approach can consistently outperform them by a notable margin on all benchmark datasets. To be more specific, enhanced VCC almost conquers UCSDped2 dataset with $99.0\%$ frame-level AUC and $4.4\%$ frame-level EER, which typically leads existing VAD methods by an over $2\%$ AUC advantage. Meanwhile, under the more strict pixel-level criteria, VCC exhibits even more obvious advantage by $96.8\%$ pixel-level AUC and $6.4\%$ pixel-level EER. When it comes to Avenue and ShanghaiTech, which are recent datasets that are more challenging, enhanced VCC also realizes a significant performance leap: While the frame-level AUC performance of existing VAD methods are basically below $90\%$  on Avenue, our VCC attains a remarkable frame-level AUC of $92.2\%$. The case is the same for ShanghaiTech, as VCC is the only method that achieves higher than $80\%$ frame-level AUC on ShanghaiTech ($80.2\%$) among various compared methods. A similar trend is also observed on EER, as VCC reports the best EER on both Avenue and ShanghaiTech. \textbf{(2)} Although the basic VCC \cite{yu2020cloze} has achieved fairly satisfactory VAD performance, enhanced VCC in this work still produces a significant improvement ($1.7\%$-$5.4\%$ AUC gain and $1.6\%$-$4.5\%$ EER reduction) when compared with its earlier counterpart. Such improvement demonstrates the sound extendibility of the proposed VCC framework, and its potential could be further exploited.

\subsubsection{Classic Datasets}

\begin{table}
	\centering
	\caption{Performance comparison on on UMN.}
	\linespread{1}\selectfont
	\scalebox{0.9}{
		\begin{tabular}{ccccc}
			\toprule
			Method & Scene1 & Scene2 & Scene3 & Average \\
			\midrule
			SF \cite{Mehran2009Abnormal} & - & - & - & 96.0\% \\
			SR \cite{cong2011sparse} & 99.5\% & 97.5\% & 96.4\% & 97.8\% \\
			LSA \cite{2012Video} & - & - & - & 98.5\% \\
			GPR \cite{hasan2016learning} & 24.0\% & 27.9\% &  &  \\
			MAC \cite{2016Combining} & 99.3\% & 96.9\% & 98.8\%  & 98.3\% \\
			SCD \cite{del2016discriminative} & - & - & - & 91.0\%  \\
			GAS \cite{2016Online} & 99.8\% & 99.3\% & 99.9\% & 99.7\%  \\
			Unmasking \cite{tudor2017unmasking} & 99.3\% & 87.7\% & 98.2\%  & 95.1\% \\
			AM-GAN \cite{ravanbakhsh2017abnormal} & - & - & - & 99.0\% \\
			TCP \cite{Ravanbakhsh2018PlugandPlayCF} & - & - & - & 98.8\% \\
			AnomalyNet \cite{zhou2019anomalynet} & - & - & - & 99.6\% \\
			NNC \cite{Ionescu2019DetectingAE} & 99.9\% & 98.2\% & 99.8\% & 99.3\% \\
			\hline
			VCC & 100.0\% & 98.9\% & 99.2\% & 99.4\% \\
			\bottomrule
	    \end{tabular}}
	\label{tab:UMN}
\end{table}

We additionally conduct experiments on three classic VAD datasets: UCSDped1, UMN and Subway exit. They are less popular in recent DNN based VAD for some intrinsic limitations: UCSDped1 suffers from low-resolution gray-scale video frames, while the height of its smallest foreground objects could be only 8-10 pixels. This is unfavorable for DNN's representation learning, while classic feature descriptor even works better; UMN contains only staged abnormal events, which means that the normal and abnormal events can be divided into non-overlapping stages, which cannot fully reflect VAD performance; Subway dataset also contains relatively blurred gray-scale frames collected by obsolete devices, and the video events are very sparse when compared with other datasets. Besides, both UMN and Subway Exit lack a separated test set and suffer from coarse annotations. Despite those limitations, we still test the enhanced VCC on them to offer a more comprehensive evaluation. Since the performance on those classic datasets are often not reported by recent DNN based VAD methods, we also include results from classic VAD methods for a reference, and the detailed results are presented in Table \ref{tab:ped1}-\ref{tab:exit}. From those results, we note that our VCC is able to achieve satisfactory performance in comparison, even though those datasets are somewhat unsuitable for DNN based VAD: On UCSDped1 dataset, we note that the proposed VCC obtains fairly competitive performance under frame-level criteria ($87.7\%$ AUC and $20.5\%$ EER). Meanwhile, under the more strict pixel-level criteria, VCC is the best performer ($76.3\%$ AUC and $27.9\%$ EER) with an evident advantage (typically $6\%$-$10\%$ AUC gain) over the compared methods. On the relatively simple UMN dataset, like most VAD methods, VCC achieves near-perfect performance under all scenes, and produces an average frame-level AUC above $99\%$; Although the sparsity of foreground on Subway Exit makes it hard for VCC to extract video events, VCC still yields a decent performance (over $91\%$ frame-level AUC), which is readily comparable to existing methods. 

\begin{table}
	\centering
	\caption{Performance comparison on Subway Exit.}
	\linespread{1}\selectfont
	\scalebox{0.9}{
		\begin{tabular}{cc}
			\toprule
			Method & AUC \\
			\midrule
			MDT \cite{mahadevan2010anomaly} & 89.7\% \\
			SR  \cite{cong2011sparse} & 80.2\% \\
			LSA \cite{2012Video} & 88.4\% \\
 			AMDN \cite{xu2017detecting} & 87.9\% \\
 			CAE \cite{hasan2016learning} & 80.7\% \\
 			SCD \cite{del2016discriminative} & 82.4\% \\
 			Unmasking \cite{tudor2017unmasking} & 85.7\% \\
 			ConvLSTM-AE \cite{luo2017remembering} & 87.7\% \\
 			FCN \cite{2018Deep} & 90.2\% \\
 			NNC \cite{Ionescu2019DetectingAE} & 95.1\% \\
 			SRNN-AE \cite{Luo2021VideoAD} & 89.7\% \\
 			SIGNet \cite{Fang2020AnomalyDW} & 95.7\% \\
 			DeepOC \cite{Wu2020ADO} & 89.5\% \\
 			\hline
 			VCC & 91.4\% \\
			\bottomrule
	    \end{tabular}}
	\label{tab:exit}
\end{table}
As a consequence, the comprehensive evaluations on different benchmark datasets have verified that our VCC is able to achieve state-of-the-art VAD performance. 

\subsection{Detailed Analysis}

\subsubsection{Ablation Studies}

For the proposed VCC framework, RoI extraction in video event extraction and two ensemble strategies in VCTs are the core to ensure superior VAD performance. To demonstrate their effectiveness, we conduct corresponding ablation studies with the basic VCC framework under the frame-level criteria. We show the results in Table \ref{tab:ablation} and analyze them accordingly: \textbf{(1)} We compare four practices to localize RoIs of foreground objects for video event extraction: Learning on a per-frame basis (FR, \textit{i.e.} no localization at all), multi-scale sliding windows with motion filtering (SDW), appearance based RoI extraction only (APR, \textit{i.e.} using a pre-trained object detector only) and the proposed method that leverages both appearance and motion cues (APR+MT). Note that the SDW's results on ShanghaiTech is not reported, since it produces excessive RoIs that are beyond the capacity of our hardware (In fact, the explosive number of STCs is exactly an important drawback of SDW). We can readily come to several conclusions: First, as can be seen from row 1, 2, 3, 6 of each dataset in Table \ref{tab:ablation}, the proposed APR+MT constantly outperforms other methods by a sensible margin. Concretely, APR+MT exhibits an obvious advantage ($2.7\%$-$4.6\%$ AUC gain) over FR and SDW, which are popular strategies adopted in VAD literature. In particular, we note that SDW performs worse than FR on both UCSDped2 and Avenue. This suggests that an imprecise localization of RoIs is even detrimental to VAD performance, so it in turn justifies the necessity of a precise localization. Meanwhile, when compared with APR, the proposed APR+MT brings evident AUC gain by $1.8\%$, $2.5\%$ and $1.2\%$ on UCSDped2, Avenue and ShanghaiTech respectively. Such observations demonstrate the importance of a more comprehensive RoI localization, which benefits VAD performance. \textbf{(2)} We compare three configurations of ensemble strategies in VCC: Without any VCT type ensemble, without modality ensemble ($w_m=0$), with both VCT type (for both appearance and motion completion) and modality ensemble. By comparing row 4, 5, 6 of each dataset in Table \ref{tab:ablation}, we can obtain two conclusions: First, VCT type ensemble boosts the VAD performance by $1.3\%$, $2.1\%$ and $0.4\%$ AUC on UCSDped2, Avenue and ShanghaiTech respectively, which justifies the need to fully exploit temporal context of video events. Second, introducing motion information by modality ensemble constantly gives better performance. Specifically, modality ensemble produces a remarkable AUC gain (up to $8\%$) on UCSDped2, while more than $1\%$ AUC improvement is also achieved on Avenue and ShanghaiTech. The great performance leap on UCSDped2 can be attributed to the fact that its gray-scale frames contain limited appearance information, while motion plays a more important role in discriminating anomalies. 

\begin{table}
	\centering
	\caption{Ablation Studies for Basic VCC.}
	\linespread{1}\selectfont
	\scalebox{0.7}{
		\begin{tabular}{cccccccc}
			\toprule
			\multirow{2}{*}{Dataset} & \multicolumn{4}{c}{Video Event Extraction} & \multicolumn{2}{c}{Ensemble} & \multirow{2}{*}{AUC}\\
			& FR & SDW & APR & APR+MT & VCT Type & Modality & \\
			\midrule
			\multirow{6}{*}{\rotatebox{90}{UCSDped2}} & \CheckmarkBold & & & & \CheckmarkBold & \CheckmarkBold & 94.6\% \\
			& & \CheckmarkBold & & & \CheckmarkBold &\CheckmarkBold & 93.3\% \\
			& & & \CheckmarkBold & & \CheckmarkBold &\CheckmarkBold & 95.5\% \\
			& & & & \CheckmarkBold &  & \CheckmarkBold & 96.0\% \\
			& & & & \CheckmarkBold & \CheckmarkBold &  & 89.6\% \\
			& & & & \CheckmarkBold & \CheckmarkBold & \CheckmarkBold & \textbf{97.3\%} \\
			\hline
			\multirow{6}{*}{\rotatebox{90}{Avenue}} & \CheckmarkBold & & & & \CheckmarkBold & \CheckmarkBold & 86.8\% \\
			& & \CheckmarkBold & & & \CheckmarkBold & \CheckmarkBold & 85.2\% \\
			& & & \CheckmarkBold & & \CheckmarkBold & \CheckmarkBold & 87.1\% \\
			& & & & \CheckmarkBold &  & \CheckmarkBold & 87.5\% \\
			& & & & \CheckmarkBold & \CheckmarkBold &  & 88.2\% \\
			& & & & \CheckmarkBold & \CheckmarkBold & \CheckmarkBold & \textbf{89.6\%} \\
			\hline
			\multirow{6}{*}{\rotatebox{90}{ShanghaiTech}} & \CheckmarkBold & & & & \CheckmarkBold & \CheckmarkBold & 70.2\% \\
			& & \CheckmarkBold & & & \CheckmarkBold & \CheckmarkBold & - \\
			& & & \CheckmarkBold & & \CheckmarkBold & \CheckmarkBold &  73.6\%\\
			& & & & \CheckmarkBold &  & \CheckmarkBold & 74.4\% \\
			& & & & \CheckmarkBold & \CheckmarkBold &  &  73.5\%\\
			& & & & \CheckmarkBold & \CheckmarkBold & \CheckmarkBold & \textbf{74.8\%} \\
			\bottomrule
	\end{tabular}}
	\label{tab:ablation}
\end{table}

\begin{table*}[ht]
	\centering
	\caption{Comparison between enhanced VCC and basic VCC.}
	\linespread{1}\selectfont
	\scalebox{0.8}{
		\begin{tabular}{cccccccccc}
			\toprule
			\multirow{2}{*}{Dataset} & \multicolumn{3}{c}{Video Event Processing} & \multicolumn{2}{c}{Network Architecture} & \multicolumn{2}{c}{Score Metric} & \multicolumn{1}{c}{Post-processing} & \multirow{2}{*}{AUC}\\
			& GD & OF & Block & UNet & ST-UNet & MSE & Mixed & Score Rectification & \\
			\midrule
			\multirow{6}{*}{\rotatebox{90}{UCSDped2}} & \CheckmarkBold & & & \CheckmarkBold &  & \CheckmarkBold & & & 97.3\% \\
			& & \CheckmarkBold & & \CheckmarkBold & & \CheckmarkBold & & & 97.6\%\\
			& & \CheckmarkBold & \CheckmarkBold & \CheckmarkBold & & \CheckmarkBold & & & 97.6\%\\
			& & \CheckmarkBold & \CheckmarkBold & & \CheckmarkBold & \CheckmarkBold & & & 98.1\% \\
			& & \CheckmarkBold & \CheckmarkBold & & \CheckmarkBold & & \CheckmarkBold & & 98.2\%\\
			& & \CheckmarkBold & \CheckmarkBold & & \CheckmarkBold & & \CheckmarkBold & \CheckmarkBold & 99.0\%\\
			\hline
			\multirow{6}{*}{\rotatebox{90}{Avenue}} & \CheckmarkBold & & & \CheckmarkBold &  & \CheckmarkBold & & & 89.6\% \\
			& & \CheckmarkBold & & \CheckmarkBold & & \CheckmarkBold & & & 90.0\%\\
			& & \CheckmarkBold & \CheckmarkBold & \CheckmarkBold & & \CheckmarkBold & & & 90.0\%\\
			& & \CheckmarkBold & \CheckmarkBold & & \CheckmarkBold & \CheckmarkBold & & & 91.7\% \\
			& & \CheckmarkBold & \CheckmarkBold & & \CheckmarkBold & & \CheckmarkBold & & 91.7\%\\
			& & \CheckmarkBold & \CheckmarkBold & & \CheckmarkBold & & \CheckmarkBold & \CheckmarkBold & 92.2\%\\
			\hline
			\multirow{6}{*}{\rotatebox{90}{ShanghaiTech}} & \CheckmarkBold & & & \CheckmarkBold &  & \CheckmarkBold & & & 74.8\% \\
			& & \CheckmarkBold & & \CheckmarkBold & & \CheckmarkBold & & & 74.8\%\\
			& & \CheckmarkBold & \CheckmarkBold & \CheckmarkBold & & \CheckmarkBold & & & 77.2\%\\
			& & \CheckmarkBold & \CheckmarkBold & & \CheckmarkBold & \CheckmarkBold & & & 77.6\% \\
			& & \CheckmarkBold & \CheckmarkBold & & \CheckmarkBold & & \CheckmarkBold & & 78.5\%\\
			& & \CheckmarkBold & \CheckmarkBold & & \CheckmarkBold & & \CheckmarkBold & \CheckmarkBold & 80.2\%\\
			\hline
			\multirow{6}{*}{\rotatebox{90}{UCSDped1}} & \CheckmarkBold & & & \CheckmarkBold & & \CheckmarkBold & & & 78.8\% \\
			& & \CheckmarkBold & & \CheckmarkBold & & \CheckmarkBold & & & 79.3\%\\
			& & \CheckmarkBold & \CheckmarkBold & \CheckmarkBold & & \CheckmarkBold & & & 87.2\%\\
			& & \CheckmarkBold & \CheckmarkBold & & \CheckmarkBold & \CheckmarkBold & & & 87.3\% \\
			& & \CheckmarkBold & \CheckmarkBold & & \CheckmarkBold & & \CheckmarkBold & & 87.3\%\\
			& & \CheckmarkBold & \CheckmarkBold & & \CheckmarkBold & & \CheckmarkBold & \CheckmarkBold & 87.7\%\\
			\hline
			\multirow{6}{*}{\rotatebox{90}{UMN}} & \CheckmarkBold & & & \CheckmarkBold & & \CheckmarkBold & & & 94.4\% \\
			& & \CheckmarkBold & & \CheckmarkBold & & \CheckmarkBold & & & 95.1\%\\
			& & \CheckmarkBold & \CheckmarkBold & \CheckmarkBold & & \CheckmarkBold & & & 97.3\%\\
			& & \CheckmarkBold & \CheckmarkBold & & \CheckmarkBold & \CheckmarkBold & & & 98.1\% \\
			& & \CheckmarkBold & \CheckmarkBold & & \CheckmarkBold & & \CheckmarkBold & & 98.1\%\\
			& & \CheckmarkBold & \CheckmarkBold & & \CheckmarkBold & & \CheckmarkBold & \CheckmarkBold & 99.4\%\\
			\hline
			\multirow{6}{*}{\rotatebox{90}{Subway Exit}} & \CheckmarkBold & & & \CheckmarkBold &  & \CheckmarkBold & & & 84.9\% \\
			& & \CheckmarkBold & & \CheckmarkBold & & \CheckmarkBold & & & 86.3\%\\
			& & \CheckmarkBold & \CheckmarkBold & \CheckmarkBold & & \CheckmarkBold & & & 86.3\%\\
			& & \CheckmarkBold & \CheckmarkBold & & \CheckmarkBold & \CheckmarkBold & & & 86.8\% \\
			& & \CheckmarkBold & \CheckmarkBold & & \CheckmarkBold & & \CheckmarkBold & & 87.0\%\\
			& & \CheckmarkBold & \CheckmarkBold & & \CheckmarkBold & & \CheckmarkBold & \CheckmarkBold & 91.4\%\\
			\bottomrule
	\end{tabular}}
	\label{tab:extension_ablation}
\end{table*}

\subsubsection{Basic VCC v.s. Enhanced VCC}

\begin{table}
	\centering
	\caption{Comparison of score rectification strategies.}
	\linespread{1}\selectfont
	\scalebox{0.9}{
		\begin{tabular}{cccc}
			\toprule
			 & UCSDped2 & Avenue & ShanghaiTech \\
			\midrule
			No Rectification & 98.2\% & 91.7\% & 78.5\% \\
			Weight Decay & 98.9\% & 92.2\% & 79.5\% \\
			Average & 99.0\% & 92.2\% & 80.2 \% \\
			Gaussian & 99.0\% & 92.1\% & 79.1\% \\
			Median & 99.1\% & 92.1\% & 80.8\% \\ 
			\bottomrule
	\end{tabular}}
	\label{tab:score_rectification}
\end{table}

As we have illustrated in Sec. \ref{sec:comp_stoa}, enhanced VCC can outperform the basic VCC by a large margin. In this section, we will provide a more detailed comparison between the basic VCC and the enhanced VCC, so as to show how enhanced VCC improves VAD performance by its every new component. With results shown in Table \ref{tab:extension_ablation}, the comparison is conducted in terms of the four aspects below: \textbf{(1)} Video event processing. The main differences in this aspect are that enhanced VCC introduces optical flow (OF) as motion cues to replace the temporal gradients (GD) for basic VCC, and it adopts a spatially-localized strategy that processes video events on a block-basis (Block). As shown by row 1 and 2 of each dataset in Table \ref{tab:extension_ablation}, using OF can constantly promote VAD performance ($0.3\%$ to $1.4\%$ AUC) on all datasets. When it comes to the spatially-localized strategy (compared by row 2 and 3), it is noted that block based processing enables significant performance improvement, especially for those datasets that are evidently influenced by foreground depth and varied foreground scales. For example, it produces $7.9\%$ and $2.4\%$ AUC improvement on UCSDped1 and ShanghaiTech dataset respectively. \textbf{(2)} Network architecture of DNN model. In this regard, enhanced VCC leverages the new ST-UNet as the network architecture of DNN model to complete VCTs, while basic VCC simply uses the standard UNet architecture. We compare them by row 3 and 4 of each dataset in Table \ref{tab:extension_ablation}: Among all benchmark datasets, we are also able to observe a consistent VAD performance enhancement brought by utilizing ST-UNet, which is up to $1.7\%$ AUC. This justifies the effectiveness to improve network architecture, which facilitates exploiting semantics and temporal context on the model level. \textbf{(3)} Score metric. We propose to use a mixed anomaly score metric that combines MSE and SSIM for enhanced VCC, which is compared with the original MSE by row 4 and 5 of each dataset in Table \ref{tab:extension_ablation}. The results show that the mixed score metric achieves comparable or superior VAD performance. In particular, an approximately $1\%$ AUC improvement is observed on ShanghaiTech, which is the most challenging benchmark with abundant high-level structure in foreground. \textbf{(4)} Score rectification. In contrast to basic VCC that ignores the post-processing step, enhanced VCC adopts temporal moving average based score rectification, which is a simple but highly effective post-processing technique. As shown in row 5 and 6 of each dataset in Table \ref{tab:extension_ablation}, we can constantly observe an unanimous and notable performance improvement ($0.5\%$-$4.4\%$ AUC gain). In addition, We also compare several different ways to set the weight $\omega_m$: Weight decay ($q=0.8$), temporal average, Gaussian ($\sigma=1$) and median. As it is shown in Table \ref{tab:score_rectification}, all types of score rectification produce better VAD performance, while median yields the best overall VAD performance. However, since computing median is more expensive, we simply choose the simplest temporal average strategy for score rectification. Consequently, score rectification is a pivotal step, and more sophisticated post-processing strategy can be further explored.  
\begin{figure*}[ht]
	\centering
	\includegraphics[scale=0.65]{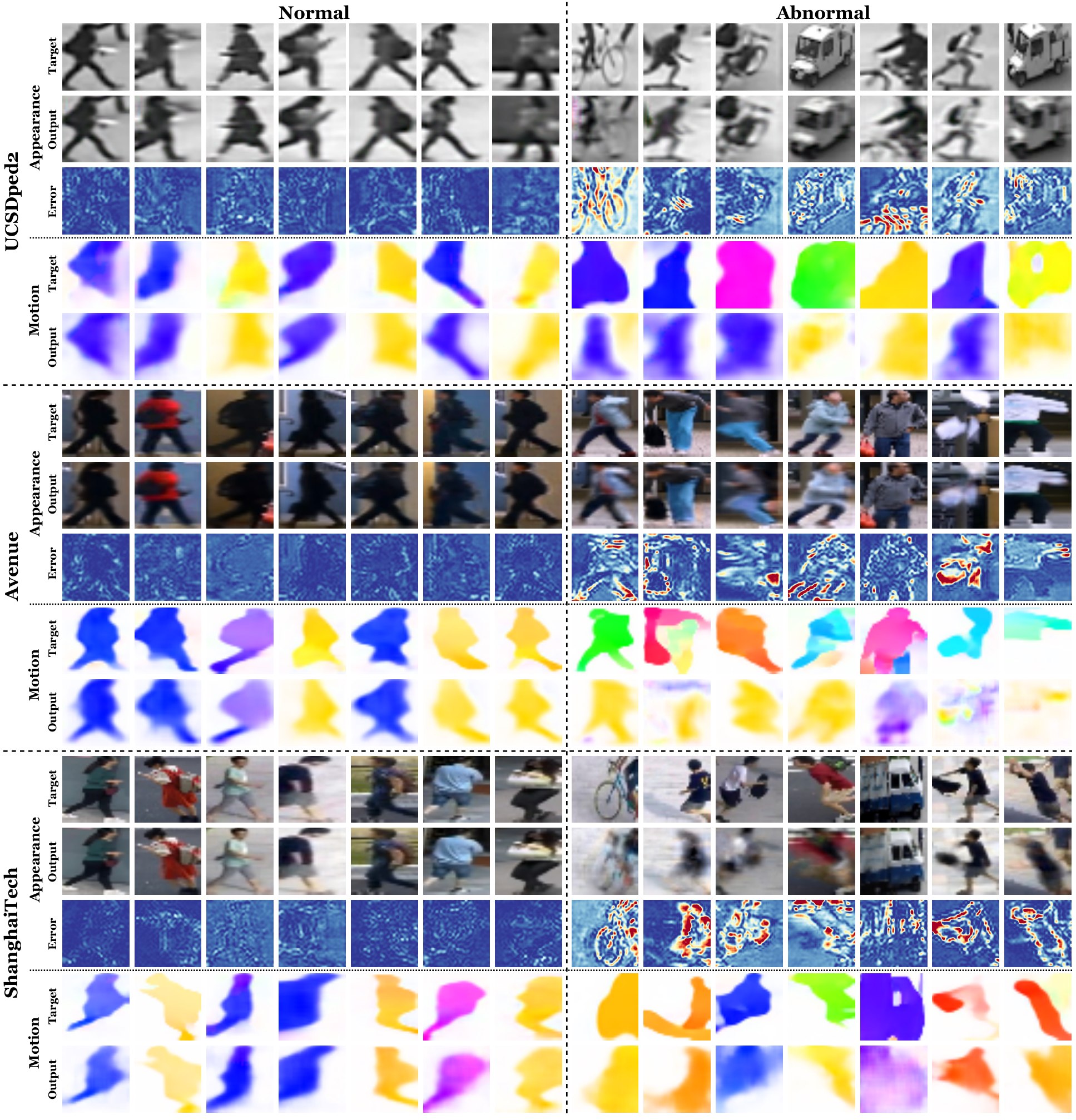}
	\caption{Visualization of erased patches and their optical flow (Target), corresponding completed patches (Output) by VCC and appearance completion error maps (Error). In the error map brighter color indicates larger error.}
	\label{fig:visual}
\end{figure*}

\subsubsection{Visualization} 

In order to reveal how VCC helps discriminate anomalies in a more intuitive manner, we visualize its generated patches and optical flow of representative normal/abnormal video events in Fig. \ref{fig:visual}. Heat maps are used to describe pixel-wise completion errors with different magnitude. Given the visualization results in Fig. \ref{fig:visual}, we discover several interesting phenomena: \textbf{(1)} For normal video events, VCC can effectively infer the erased patches and their optical flow. In this case, most completion errors are moderate. At the same time, the optical flow of erased patches can also be soundly restored. By contrast, abnormal video events incurs sharp and prominent errors in completion, while their motion completion also suffers from obvious errors in both magnitude and direction of optical flow. \textbf{(2)} It can be observed that the completion errors of normal events are distributed around foreground object contour in a relatively uniform manner. By comparison, the distribution of abnormal events' completion errors is highly non-uniform, and most of them possess a straightforward semantic interpretation. As displayed by heat maps, intense completion errors of abnormal events are often concentrated on some high-level parts of foreground objects, \textit{e.g.} the riding bicycle, falling paper with wrongly inferred shape and position, lifted or thrown backpack, body parts like legs and hands. By contrast, other regions are endowed with smaller errors.

\subsubsection{Additional Remarks} 

\begin{table}
	\centering
	\caption{Average per-frame computation cost of VCC.}
	\linespread{1}\selectfont
	\scalebox{0.9}{
		\begin{tabular}{cccc}
			\toprule
			Dataset & Video Event Extraction & Inference & Total\\
			\midrule
			UCSDped2 & 0.11s & 0.05s & 0.16s\\
			Avenue & 0.15s & 0.04s & 0.19s\\
			ShanghaiTech & 0.19s & 0.02s & 0.21s\\
			\bottomrule
	\end{tabular}}
	\label{tab:computation_cost}
\end{table}

Apart from the discussion in previous sections, we would like to make some additional remarks on the proposed VCC approach: \textbf{(1)} Connections to frame prediction and reconstruction. In fact, when the whole frame is considered as one video event and only type-$D$ VCTs are completed, VCC is equivalent to frame prediction. In other words, frame prediction can be viewed a special case of VCC, while VCC enjoys superior VAD performance to the case using only type-$D$ VCTs (see Table \ref{tab:ablation}) and frame prediction based methods (see Table \ref{tab:p2_av_sh}). Besides, when the core component of VCC, \textit{i.e.} VCTs, are replaced by plain reconstruction of STCs, our experiments usually report a $3\%$ to $7\%$ AUC loss, which validates the necessity of VCTs. \textbf{(2)} The computation cost of VCC. As shown in Table \ref{tab:computation_cost}, for the proposed enhanced VCC, it takes 0.16s, 0.19s and 0.21s on average to process one frame on UCSDped2, Avenue and ShanghaiTech dataset respectively, which is fairly acceptable under a Python implementation. Meanwhile, it is easy to note that video event extraction is actually the computational bottleneck that costs the most time. To speed up video event extraction, more efficient pre-trained object detector and optical flow estimator can be leveraged. Besides, since each ST-UNet model is independent, it is convenient to parallelize the inference and realize further acceleration.

\section{Conclusion}
\label{sec:conclusion}
This paper proposes VCC as a new solution to DNN based VAD. VCC first utilizes appearance and motion as complimentary cues to extract RoIs of foreground objects, so as to accomplish both precise and comprehensive video event extraction. By erasing a certain patch, each video event is transformed into a VCT. To solve a VCT, DNNs are trained to infer the erased patch and its optical flow, which spurs DNNs to capture video semantics rather than low-level details. Subsequently, VCC is then equipped with VCT type ensemble and modality ensemble, which enable VCC to fully exploit spatio-temporal context and consider richer motion information. To further ameliorate VCC, we develop a series of practical strategies and propose the enhanced VCC. Extensive empirical evaluations justify VCC as a highly effective VAD solution that achieves state-of-the-art performance under both frame-level and pixel-level criteria. For our future research, considering the effectiveness of video event extraction in VCC, it would be meaningful to explore a general strategy that can realize both precise and comprehensive RoI localization with a non-fixed camera, \textit{e.g.} using an open-world object detector. Second, we intend to explore the VCC as a potential solution for weakly-supervised and unsupervised VAD. 

\backmatter








\section*{Declarations}


The work is supported by National Natural Science Foundation of China under Grant No. 62006236, Hunan Provincial Natural Science Foundation under Grant No. 2020JJ5673, and NUDT Research Project under Grant No. ZK20-10. The authors are employed at National University of Defense Technology/Dongguan University of Technology. All authors agree with the content of this work and give explicit consent to submitting and publishing this work. All datasets and codes used in this work are publicly available. The codes are available at \url{https://github.com/yuguangnudt/VEC_VAD/tree/VCC}.

\noindent






\bibliographystyle{sn-chicago}
\bibliography{sn-bibliography}



\begin{appendices}

\section{}

\subsection{Proof of Theorem 1}\label{secA1}

\textbf{Proof:} Consider a $n$-dimensional space, where $n\in \mathcal{N}^+$ is a positive integer. The $n$-dimensional space is uniformly partitioned into $n$-dimensional local hyper-cube regions $\{R^n_i\}^{\infty}_{i=1}$ with any size. Suppose that an arbitrary $n$-dimensional hyper-cube $b^n$ is overlapped with some $R^n_i$. Since both $b^n$ and $R^n_i$ are hyper-cubes, their overlap region is still a hyper-cube. Thus, it is straightforward for us to define a generalized ``volume'' of the overlap region by:

\begin{equation}
\label{eq:vol}
    \mathcal{O}(b^n, R^n_i)=\prod^n_{j=1} m^{(j)}_i
\end{equation}
where $m^{(j)}_i$ denotes the length of the overlap region between $b^n$ and $R^n_i$ on the $j_{th}$ dimension. Assume that $b^n$'s geometric center $c^n_b=[c_1, c_2,\cdots, c_n]$ falls into the $k_{th}$ region $R^n_k$, where $c_j$ is the coordinate of $c^n_b$ on the $j_{th}$ dimension. Suppose that $R^n_i$ is located at the range $(l^{(j)}_i, u^{(j)}_i)$ on the $j_{th}$ dimension, where $l^{(j)}_i<u^{(j)}_i$. Since $c^n_b$ falls into $R^n_k$, we can know that:
\begin{equation}
    l^{(j)}_k<c_j<u^{(j)}_k
\end{equation}
We can simply consider the 1-dimensional case on the $j_{th}$ dimension (see Fig. \ref{fig:discussion}). Suppose that the edge of hyper-cube $b^n$ on $j_{th}$ dimension is the line segment $b^n_j$, whose center point is $c_j$. A line segment that starts at $l^{(j)}_i$ and ends at $u^{(j)}_i$ is denoted by $L_i$, which also represents the range of region $R^n_k$ on the $j_{th}$ dimension. Naturally,  $m^{(j)}_i$ is equivalent to the overlap length between $b^n_j$ and $L_i$, and we have:

\begin{figure}[ht]
	\centering
	\includegraphics[scale=0.5]{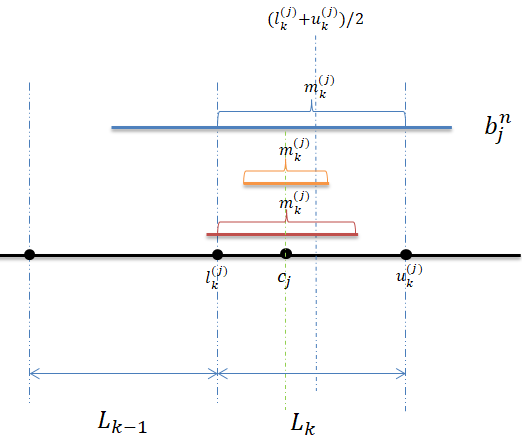}
	\caption{Three possible cases for the overlap of $b^n_k$ and $L_k$.}
	\label{fig:discussion}
\end{figure}

\begin{equation}
    0\leq m^{(j)}_{i}\leq \min\{\vert L_i\vert, \vert b^n_j\vert\}, \quad i=1,2,\cdots
\end{equation}
where $\vert\cdot \vert$ denotes the length of a line segment. To further facilitate analysis, we can assume that $c_j$ lies at the left half of $L_k$, \textit{i.e.} $ l^{(j)}_k<c_j<\frac{u^{(j)}_k+l^{(j)}_k}{2}$, because the other case can be analyzed similarly by symmetry. We define $L_{i-1}$ to be the adjacent line segment that is on the left side of $L_i$. Then, our goal is to prove that: 

\begin{equation}
\label{eq:1d}
    m^{(j)}_{k} = \sup_i\{m^{(j)}_{i}\}
\end{equation}
For proof, we can discuss the value of $m^{(j)}_{k}$ in terms of three different cases, which are visualized by line segments with three different colors in Fig. \ref{fig:discussion}: \textbf{(1)} The entire $L_k$ is overlapped with $b^n_j$ (blue line). In this case, it is easy to know $ m^{(j)}_{k}=\vert L_k\vert$. Since $L_k=L_i$ (uniform partition) and $
m^{(j)}_{i}\leq\min\{\vert L_i\vert, \vert b^n_j\vert\}\leq\vert L_k\vert= m^{(j)}_{k}$, we have $m^{(j)}_{k} = \sup_i\{m^{(j)}_{i}\}$.  \textbf{(2)}  The entire $b^n_j$ is overlapped with $L_k$ (orange line). In this case, it is easy to know $ m^{(j)}_{k}=\vert b^n_j\vert$. Since $
m^{(j)}_{i}\leq\min\{\vert L_i\vert, \vert b^n_j\vert\}\leq\vert b^n_j\vert= m^{(j)}_{k}$, we have $m^{(j)}_{k} = \sup_i\{m^{(j)}_{i}\}$. \textbf{(3)} A portion of $L_k$ is overlapped with a portion of $b^n_j$ (red line). In this case, $b^n_j$ is only overlapped with $L_k$ and $L_{k-1}$. Since the center of $b^n_j$ is on the side of $L_k$, it is obvious that $m^{(j)}_{k}>m^{(j)}_{k-1}$. Since $m^{(j)}_{i}=0$ when $i$ is not $k$ or $k-1$, we have $m^{(j)}_{k} = \sup_i\{m^{(j)}_{i}\}$. In this way, we show that Eq. \ref{eq:1d} always holds. As a consequence, we have:

\begin{equation}
\prod^n_{j=1} m^{(j)}_k=\prod^n_{j=1}\sup_i\{m^{(j)}_{i}\}=\sup_i\{\prod^n_{j=1}m^{(j)}_{i}\}
\end{equation}
Given the definition of volume in Eq. \ref{eq:vol}, we have:

\begin{equation}
    \mathcal{O}(b^n, R^n_k)=\sup_i\{\mathcal{O}(b^n, R^n_i)\}
\end{equation}
Consequently, Theorem 1 in the manuscript holds.

\subsection{Implementation Details} \label{secA2}
We report the implementation details of enhanced VCC in this section, while the parameterization of basic VCC is provided in our previous work \cite{yu2020cloze}. For video event extraction, cascade R-CNN \cite{cai2018cascade} pre-trained on Microsoft COCO dataset is used as object detector as it achieves a good trade-off between performance and speed. As to STCs, we set $h=w=32$ and $D=5$. Meanwhile, we set confidence score threshold $T_s=0.5$, optical flow binarization threshold $T_b=1$ and maximum aspect-ratio threshold $T_{ar}=10$ for all benchmark datasets in our experiments. Considering the size of video frames and foreground objects in different datasets, RoI area threshold $T_a$ is set to $12\times12$ for UCSDped1/UCSDped2, $40\times40$ for Avenue and Subway Exit, $30\times30$ for ShanghaiTech, and $10\times10$ for UMN respectively. The overlapping ratio $T_o$ is set to 0.6 for UCSDped1/UCSDped2 and Avenue, while $T_o=0.65$ is adopted for ShanghaiTech, Subway Exit and UMN. To handle benchmark datasets with varied foreground depth (UCSDped1, ShanghaiTech and UMN Scene2/Scene3), we uniformly divide their video frame into $4\times 1$, $4\times 1$ and $2\times 1$ blocks respectively. For other datasets without evident depth variation, we simply impose no block partition, which is equivalent to a $1\times 1$ block. To perform VCC, we adopt the proposed ST-UNet as basic DNN architecture, which is optimized by an Adam optimizer with a learning rate $lr=0.002$ for ShanghaiTech and $lr=0.001$ for other datasets. Each ST-UNet is trained by 5 epochs for UCSDped2, 10 epochs for Avenue and ShanghaiTech, 20 epochs for UMN, and 30 epochs for UCSDped1 and Subway Exit. Besides, the batch size is set to 200 for ShanghaiTech and 128 for other datasets. For modality ensemble, we set $(w_a, w_m)$ to be $(0.5,1)$ for UCSDped2 and Subway Exit, $(1,1)$ for Avenue, UCSDped1 and UMN Scene1, and $(1,0.5)$ for ShanghaiTech, UMN Scene2/Scene3 respectively. As to the mixed score metric, the weight of MSE is fixed to 1, while SSIM is usually weighted by 1 or 0.1. For score rectification, the window size $W$ is set to 5 for UCSDped1/UCSDped2 and Avenue, 20 for ShanghaiTech, 50 for UMN, and 200 for Subway Exit respectively. The entire VCC approach is implemented with Pytorch framework \cite{paszke2019pytorch} under Python 3.6 programming environment, and all of our experiments are carried out on a PC with 128 GiB RAM, Nvidia Titan Xp GPUs and a 3.7GHz Intel i9-10900X CPU.





\end{appendices}

\end{document}